\documentclass[letterpaper, 10 pt, journal, twoside]{IEEEtran}

\usepackage{graphics} 
\usepackage{epsfig} 
\usepackage{amsmath} 
\usepackage{amssymb}  

\usepackage{pgfplots}  
\pgfplotsset{compat=1.17}  
\usepackage{tikz}  
\usepackage{adjustbox}  
\definecolor{color1}{HTML}{a61c00}
\definecolor{color2}{HTML}{1155cc}
\definecolor{color3}{HTML}{6d9eeb}
\definecolor{color4}{HTML}{a4c2f4}

\usepackage{booktabs}
\usepackage{nicefrac}       

\usepackage{graphicx}

\newcommand{\yes}{\checkmark}
\newcommand{\no}{$\times$}
\usepackage[caption=false]{subfig}

\newcommand{\PAR}[1]{\noindent{\bf #1}}

\usepackage{multirow}
\usepackage[misc,geometry]{ifsym}
\newcommand{\Ours}{CAFuser}
\usepackage{amssymb}

\usepackage{colortbl}
\usepackage{pgf}

\usepackage[pagebackref=true,breaklinks=true,letterpaper=true,colorlinks,bookmarks=false]{hyperref}
\usepackage[capitalize]{cleveref}
\crefname{section}{Sec.}{Secs.}
\Crefname{section}{Section}{Sections}
\Crefname{table}{Table}{Tables}
\crefname{table}{Tab.}{Tabs.}

\begin{document}
\title{CAFuser: Condition-Aware Multimodal Fusion\\
for Robust Semantic Perception of Driving Scenes}

\author{Tim Br{\"o}dermann$^{1}$, Christos Sakaridis$^{1}$, Yuqian Fu$^{1,2 }$, and Luc Van Gool$^{1,2}$
\thanks{© 2024 IEEE.  Personal use of this material is permitted.  Permission from IEEE must be obtained for all other uses, in any current or future media, including reprinting/republishing this material for advertising or promotional purposes, creating new collective works, for resale or redistribution to servers or lists, or reuse of any copyrighted component of this work in other works.
This work was supported by the ETH Future Computing Laboratory (EFCL), financed by a donation from Huawei Technologies.
\textit{(Corresponding authors: Tim Br{\"o}dermann; Yuqian Fu.)}
}%
\thanks{$^{1}$Tim Br{\"o}dermann, Christos Sakaridis, Yuqian Fu, and Luc Van Gool are with Computer Vision Laboratory, ETH Zurich, 8057 Zurich, Switzerland  {\tt\footnotesize\{timbr, csakarid, vangool\}@vision.ee.ethz.ch}.} %
\thanks{$^{2}$Yuqian Fu and Luc Van Gool are with INSAIT, Sofia University St.\ Kliment Ohridski, Bulgaria, {\tt\footnotesize\{yuqian.fu, luc.vangool\}@insait.ai}.}%
}

\markboth{IEEE Robotics and Automation Letters. Preprint Version. Accepted January, 2025}
{Br{\"o}dermann \MakeLowercase{\textit{et al.}}: CAFuser: Condition-Aware Multimodal Fusion}

\maketitle

\begin{abstract}
    Leveraging multiple sensors is crucial for robust semantic perception in autonomous driving, as each sensor type has complementary strengths and weaknesses. However, existing sensor fusion methods often treat sensors uniformly across all conditions, leading to suboptimal performance. By contrast, we propose a novel, \emph{condition-aware multimodal} fusion approach for robust semantic perception of driving scenes. Our method, \Ours{}, uses an RGB camera input to classify environmental conditions and generate a \emph{Condition Token} that guides the fusion of multiple sensor modalities. We further newly introduce modality-specific feature adapters to align diverse sensor inputs into a shared latent space, enabling efficient integration with a single and shared pre-trained backbone. By dynamically adapting sensor fusion based on the actual condition, our model significantly improves robustness and accuracy, especially in adverse-condition scenarios. 
    \Ours{} ranks first on the public MUSES benchmarks, achieving 59.7 PQ for multimodal panoptic and 78.2 mIoU for semantic segmentation, and also sets the new state of the art on DeLiVER. 
    The source code is publicly available at: \url{https://github.com/timbroed/CAFuser}.
\end{abstract}


\IEEEpeerreviewmaketitle

\section{Introduction}

\IEEEPARstart{C}{urrent} perception pipelines for automated driving systems yield excellent results under normal, clear-weather conditions, but still struggle when they encounter adverse conditions. This prevents achieving the ultimate Level-5 driving automation, which requires a reliable perception system with an unlimited operational design domain (ODD). A major associated challenge is the accurate pixel-level semantic parsing of driving scenes, as experimental evidence~\cite{does:vision:matter:for:action} suggests that such high-level parsing is beneficial for the downstream driving tasks of prediction and planning.

Because of the aforementioned universal ODD requirement, using a single type of sensor for dense semantic perception of driving scenes is a fragile choice. More specifically, the sensitivity patterns of different sensors across environmental conditions differ drastically. For instance, while standard, RGB frame-based cameras have excellent spatial resolution, their measurements degrade severely in low illumination and adverse weather. Lidars and event cameras are by contrast much more robust to ambient lighting, but they are also strongly affected by weather particles such as raindrops or snowflakes. Radars handle adverse weather excellently but offer a far more limited spatial resolution. Thus, utilizing inputs from all sensor modalities of a vehicle's suite in a multimodal fusion framework shows much more promise for reliable semantic perception, especially in adverse ODDs such as nighttime, fog, rain, or snowfall.

\begin{figure}[t]
  \centering
  \includegraphics[width=\linewidth,
  trim={0cm 9.5cm 14cm 0cm}    
  ]{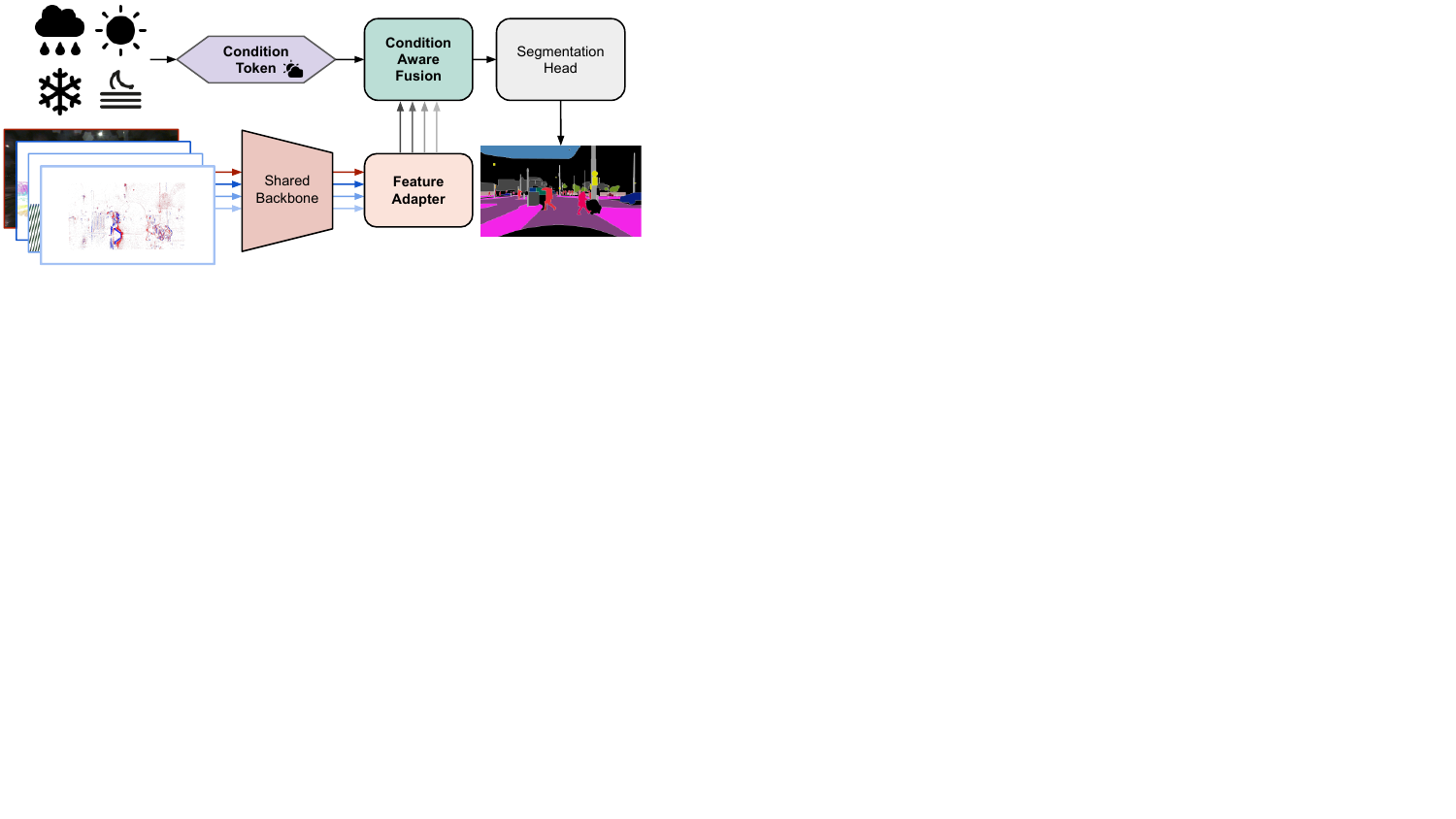}
  \caption{\textbf{CAFuser overview}. We encode the weather and lighting conditions in a Condition Token and guide the condition-aware fusion with it.
  }
  \label{fig:teaser}
\end{figure}

We observe that despite the decreasing cost of the aforementioned types of sensors and their growing adoption in autonomous vehicles, little attention has been paid to leveraging their complementary strengths in a \emph{condition-aware} manner. That is, most current multimodal fusion methods fuse sensors \emph{uniformly} across all environmental conditions. However, as the reliability of each sensor depends strongly on these conditions, fusing all sensors in a condition-agnostic fashion can generally lead to suboptimal performance. Thus, we propose a multimodal condition-aware fusion (CAF) module for robust semantic perception of driving scenes. By explicitly representing environmental conditions and adjusting the sensor fusion algorithm to these representations, we aim to arrive at an adaptive, condition-aware sensor fusion model, which knows its ODD and has learned which sensors are more informative and reliable for perception in that ODD.

Our network uses the RGB camera input to
generate a \emph{Condition Token} (CT), which guides the fusion of multiple sensors, ensuring that all sensors interact optimally. Learning this token is supervised with a verbo-visual contrastive loss utilizing text prompts \`a la CLIP~\cite{radford2021learning}, so as to ensure that the CT embeddings are aligned with the abstract language-based descriptions of environmental conditions, which can be provided as frame-level annotations of the multimodal input data~\cite{MUSES}.
By training the system end-to-end, our model learns to dynamically adapt to actual conditions, ensuring that sensor fusion is optimized for each condition, leading to more accurate semantic parsing.

Additionally, most existing fusion approaches rely on separate encoders for each sensor modality, leading to high computational complexity and requiring separate training pipelines for each modality. However, in real-time automated driving systems, efficiency is of paramount importance. In this regard, recent work~\cite{wu2023eventclip} has demonstrated the effectiveness of large-scale pre-trained models even for non-RGB-camera modalities, showing the feasibility of using a shared backbone across multiple sensor modalities. This motivates us to introduce a \emph{single network backbone} for extracting features from different sensor modalities, while still preserving the unique information from each modality.

Thus, our novel condition-aware multimodal fusion network design comprises a shared backbone for all modalities, combined with individual lightweight \emph{feature adapters}~\cite{gao2024clip} for each modality. We project all sensor inputs onto the image plane, as in prior multimodal segmentation approaches~\cite{broedermann2023hrfuser, zhang2023delivering, jia2024geminifusion}, in order to make them all compatible with the backbone they are fed to. Beyond being efficient, this design has two additional merits: 1) using the same backbone naturally allows non-RGB modalities such as lidar, radar, and event camera to get mapped to an RGB-compatible feature space, and 2) the feature adapters enable extraction of modality-specific information, providing complementary features to the RGB modality. Our experiments show a substantial reduction in model parameters (by 54\%) via this design, while not sacrificing performance.

Extensive experiments show that our method, \Ours{} (Condition-Aware Fuser), effectively learns to weigh different modalities in the fusion according to the present condition. 
\Ours{} sets the new state of the art on DeLiVER~\cite{zhang2023delivering} and ranks
first on the public MUSES~\cite{MUSES} benchmarks for both multimodal panoptic and semantic segmentation.

Our main contributions can be summarized as:
\begin{itemize}
    \item \textbf{Condition-aware fusion:} 
    We propose a \textit{Condition Token} that guides our model to adaptively fuse based on the present environmental conditions, significantly improving robustness in adverse scenarios.
    \item \textbf{Efficient architecture:}
    We newly utilize feature adapters in a shared-backbone sensor fusion architecture. Each sensor's features are aligned in a shared latent space, allowing efficient integration with the pre-trained backbone and reducing model size significantly.
    \item \textbf{Modular and scalable design:} Our architecture allows flexible and efficient addition of diverse sensor modalities, being adaptable to various sensor setups.
    \item \textbf{SOTA Performance:} We extensively ablate our method and demonstrate SOTA performance in multimodal panoptic and semantic segmentation.
\end{itemize}

\section{Related Work}

\noindent{\textbf{Semantic perception}
involves understanding the environment in a scene, encompassing tasks like semantic, instance, and panoptic segmentation.
Semantic segmentation~\cite{Cityscapes} focuses on classifying each pixel into a predefined category, while instance segmentation distinguishes between individual instances of objects within  ``things'' (e.g., cars, pedestrians) categories.
Panoptic segmentation~\cite{kirillov2019panoptic, elharrouss2021panoptic} unifies semantic and instance segmentation by predicting both ``stuff'' (e.g., sky, road) and ``things'' in an image.
These tasks are crucial for autonomous vehicles and robotics in general, as they provide the detailed environmental understanding necessary for downstream tasks such as path planning, obstacle avoidance, and applications like domestic service robotics~\cite{hurtado2022semantic}.
Recent transformer-based architectures like MaskFormer~\cite{cheng2021per} and Mask2Former~\cite{cheng2022masked} have advanced semantic perception. EfficientPS~\cite{mohan2021efficientps} improves efficiency without sacrificing accuracy, MaskDINO~\cite{li2023mask} predicts object masks simultaneously with bounding boxes, and OneFormer~\cite{jain2023oneformer} achieves state-of-the-art results in instance, semantic, and panoptic segmentation with a single architecture and model.
In contrast to these RGB-only methods, we tackle multimodal semantic perception by leveraging diverse sensor modalities for robust scene understanding in autonomous driving. Utilizing the recent MUSES dataset~\cite{MUSES}, which includes an RGB camera, a lidar, a radar, and an event camera, we enhance the reliability and accuracy of semantic perception.

\begin{figure*}[h!]
  \centering
  \includegraphics[width=\textwidth, 
  trim={0cm 1cm 6.7cm 0cm}
  ]{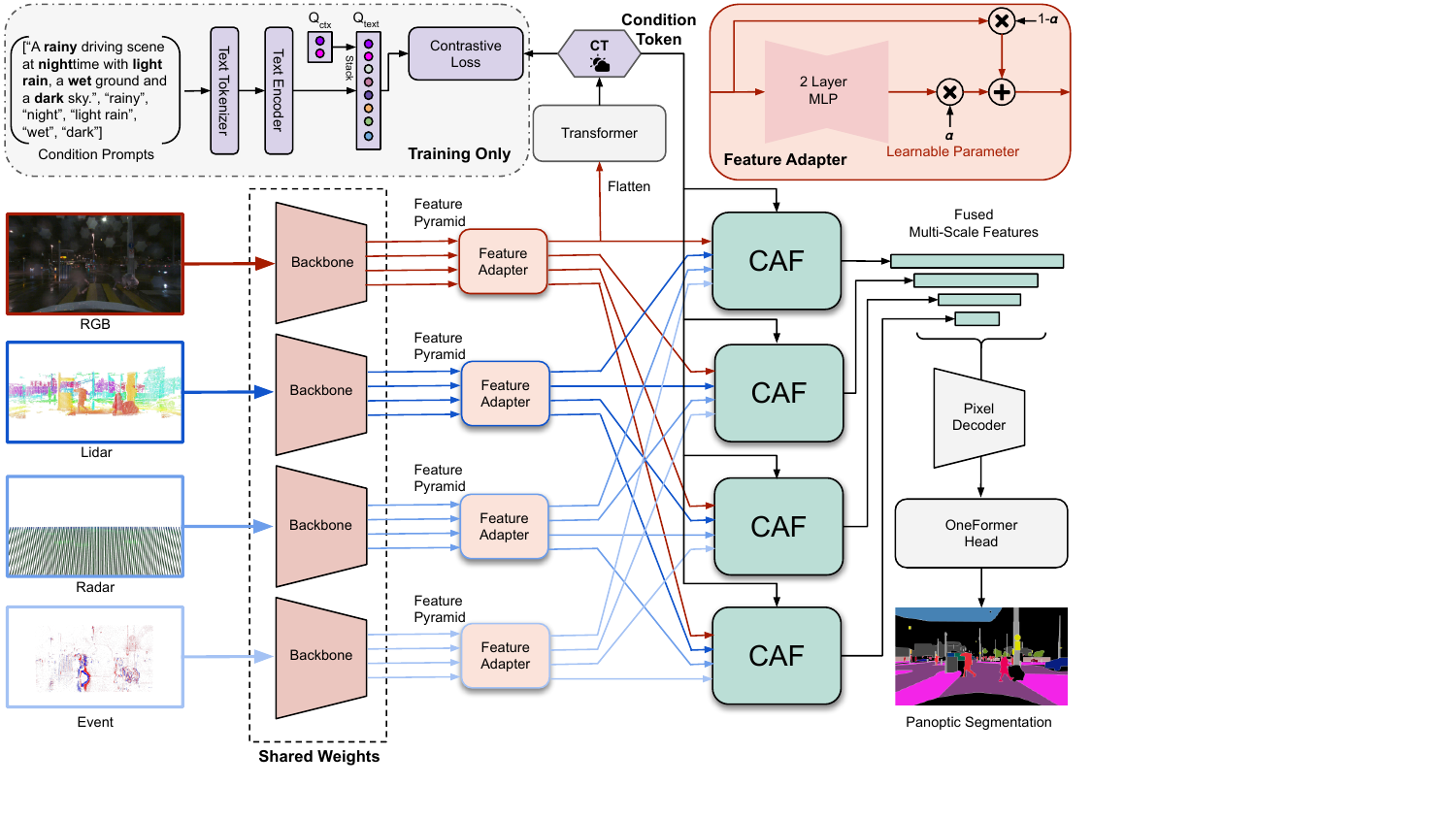} 
  \caption{\textbf{Our proposed \Ours{} architecture} with RGB camera, lidar, radar, and event camera as input modalities. Each input is passed through the shared backbone and an individual feature adapter. The CT is generated from the highest-level RGB feature map, supervised with a verbo-visual contrastive loss to our encoded condition prompts ($Q_{text}$), and used to guide the condition-aware fusion (CAF). The resulting fused multi-scale feature maps are then passed to the pixel decoder and the OneFormer~\cite{jain2023oneformer} head to produce the prediction.}
  \label{fig:overview}
\end{figure*}

\noindent{\textbf{Multimodal feature fusion} is crucial in autonomous driving, as different sensors provide complementary information. Early research focused on enhancing lidar-based 3D detection with RGB camera data~\cite{deep:continuous:fusion:3d:detection}. Large-scale datasets such as KITTI~\cite{kitti} and nuScenes~\cite{nuScenes} propelled this research but lacked recordings under adverse weather conditions, which motivated subsequent synthetic~\cite{zhang2023delivering} and real-world~\cite{seeing_through_fog, 
CADC, 
ACDC, MUSES} datasets focusing on challenging environments. 

Fusion techniques evolved from fusing two specific modalities~\cite{low_latency, chaturvedi_pay} to RGB-X fusion with arbitrary modalities~\cite{zhang2023cmx, huang2024roadformer+}. Methods like HRFuser~\cite{broedermann2023hrfuser} and CMNeXt~\cite{zhang2023delivering} introduced architectures capable of handling multiple arbitrary sensor inputs, employing modular designs and attention mechanisms. 
StitchFusion~\cite{li2024stitchfusion} integrated large-scale pre-trained models directly as encoders and feature fusers, using a multi-directional adapter module for cross-modal information transfer during encoding. Recent works such as SAMFusion~\cite{palladindietzesamfusion} explore modality-specific multimodal fusion for 3D detection, and~\cite{centeringthevalue} examines modality-agnostic multimodal fusion for semantic segmentation by employing a shared backbone but lacking feature adapters, leading to over-reliance on two modalities and no performance gain when adding lidar or event data. GeminiFusion~\cite{jia2024geminifusion} combines intra-modal and inter-modal attention to dynamically integrate complementary information across modalities, representing the state of the art in multimodal semantic segmentation. MUSES~\cite{MUSES} tackles multimodal panoptic segmentation by local-window cross-attention to merge features from multiple independent backbones.

While these works enhance perception in challenging environments, they generally fuse sensor modalities uniformly and lack explicit adaptation to environmental conditions such as fog or low light. In contrast, we perform condition-aware fusion in a shared latent space, allowing our model to adapt dynamically to environmental conditions, thereby improving the robustness of multimodal semantic perception.

\noindent{\textbf{Condition-aware perception} incorporates knowledge of the environmental conditions to guide perception. \cite{chaturvedi:pay:Attention:Weather} tackles 2D object detection by implicitly allocating higher weights to the modality with better detection features at the late-stage fusion. CoLA~\cite{cola} enhances salient object detection by leveraging pre-trained vision-language models with prompt learners to adjust for noisy or missing inputs. Knowledge distillation from vision-language models is used in~\cite{learningmodalityagnostic} for modality-agnostic representations, without focusing on condition-aware predictions. By contrast, we explicitly encode detailed environmental conditions using a Condition Token supervised by verbal scene descriptions. This token dynamically guides our sensor fusion, improving the robustness of segmentation across diverse ODDs.

\noindent{\textbf{Feature adapters}~\cite{gao2024clip, yang2024mma, chen2022vision} 
offer a lightweight solution for integrating different sensor modalities into shared models. CLIP-Adapter~\cite{gao2024clip} uses MLP adapters with residual connections to adapt features without overfitting. Modality-shared and modality-specific adapters were employed in~\cite{lu2021rgbt} for RGB-thermal tracking. EventClip~\cite{wu2023eventclip} aligns event data with CLIP features using feature adapters. While these methods focus on bimodal fusion, our network extends feature adapters to align diverse inputs from several modalities in a shared latent space, enabling flexible multimodal fusion.

\section{Method}
\label{sec:method}

We build on the recent OneFormer~\cite{jain2023oneformer}, using its head and segmentation framework as our starting point. In contrast to previous multimodal approaches, we introduce a single, shared backbone for all sensor modalities (see \cref{fig:overview}). By employing lightweight adapters for efficient feature transformation, this design significantly reduces model parameters while maintaining competitive performance.
We initialize the backbone with ImageNet pre-training, which corresponds to the RGB camera modality. For pre-processing, we follow MUSES~\cite{MUSES} and project each sensor's data (e.g., lidar, radar) as a 3-channel image onto the RGB plane. Further, we normalize each modality over the entire dataset to ensure an input representation consistent with the RGB modality.

\subsection{Multimodal Adapter}

As the feature adapter, we employ a 2-layer MLP with a 4x reduction in hidden dimensions~\cite{gao2024clip}. 
A learnable parameter $\alpha$ controls the weighting of adapted and original features. Each modality and each feature map from the Swin backbone’s 4-level feature pyramid is adapted using an individual adapter at each stage. For 4 modalities and 4 feature levels this results in 16 individual lightweight adapters. In \cref{sec:abl} we show that this setup allows us to reduce the parameters by 54\% without any loss in performance.

\subsection{Condition-Aware Fusion}
\label{sec:methods:caf}

Since sensor reliability changes predictably based on environmental conditions, we introduce a condition-aware fusion (CAF) mechanism that dynamically adapts sensor fusion in response to the current ODD. 
As labeled condition data cannot be assumed to be available at inference time, we generate a \textit{Condition Token} from the RGB camera input and use it to modulate the fusion process. The RGB camera captures sufficient global environmental information to effectively represent scene conditions, avoiding the computational overhead of processing additional modalities.

\PAR{Condition Token (CT):} Our CT generation, as shown in \cref{fig:overview}, starts by flattening the highest-level RGB feature map and passing it through a Transformer with 2 encoder and 2 decoder layers.
The resulting CT is directly supervised during training using a verbo-visual contrastive loss utilizing text prompts based on a detailed description of the environmental condition. 
For this, the MUSES dataset provides several key scene attributes, including \textit{weather condition},\textit{ precipitation type} and \textit{level}, \textit{ground condition}, \textit{time of day}, and \textit{sky condition}. We automatically create a \emph{condition prompt} from these attributes by slightly adapting them to fit into a continuous sentence. For example, the sky condition of \textit{Sunlight} becomes \textit{a sunny sky}. We further combine the precipitation type and level into one precipitation text (e.g.\ \textit{light rain}) and fill in empty condition labels from context: e.g.\ at nighttime the sky condition attribute is often missing and is filled in as \textit{dark}.
Using these attributes, we construct a rich, descriptive condition prompt for each scene using the following template:
\begin{equation}
\begin{aligned}
    \text{A } \{\text{weather condition}\} \text{ driving scene at } & \\
    \{\text{time of day}\}\text{time with } \{\text{precipitation text}\}, & \\
    \text{ a } \{\text{ground condition}\} \text{ ground and a } \{\text{sun level}\} \text{ sky.}
\end{aligned}
\end{equation}
For example, our condition prompt may be instantiated as 
``A \textit{rainy} driving scene at \textit{night}time with \textit{light rain}, a \textit{wet} ground and a \textit{dark} sky.''
This detailed prompt is combined with the individual scene attributes and guides the CT to capture the nuanced environmental context needed for robust fusion.

We follow~\cite{jain2023oneformer} to generate text queries ($Q_{text}$) from our encoded condition prompts. This step includes a tokenizer for the condition prompts and a pass through a 6-layer transformer text encoder~\cite{xu2022groupvit} and stacking with four context tokens $Q_{ctx}$~\cite{jain2023oneformer, mtaclip}.
For the resulting $Q_{text}$ and the CT, we apply a $CT-Q_{text}$ contrastive loss~\cite{jain2023oneformer,bruggemann2023contrastive,xu2022groupvit}.

At inference, we dynamically generate the CT directly from the RGB input, removing any reliance on explicit condition classification. This approach allows the CT to regulate the fusion on-the-fly based on the current environmental context, without any condition labels.
Using the CT, we explore two fusion strategies:

\begin{figure}[tb]
  \centering
  \includegraphics[width=\linewidth,
  trim={0cm 5.7cm 12.7cm 0cm}    
  ]{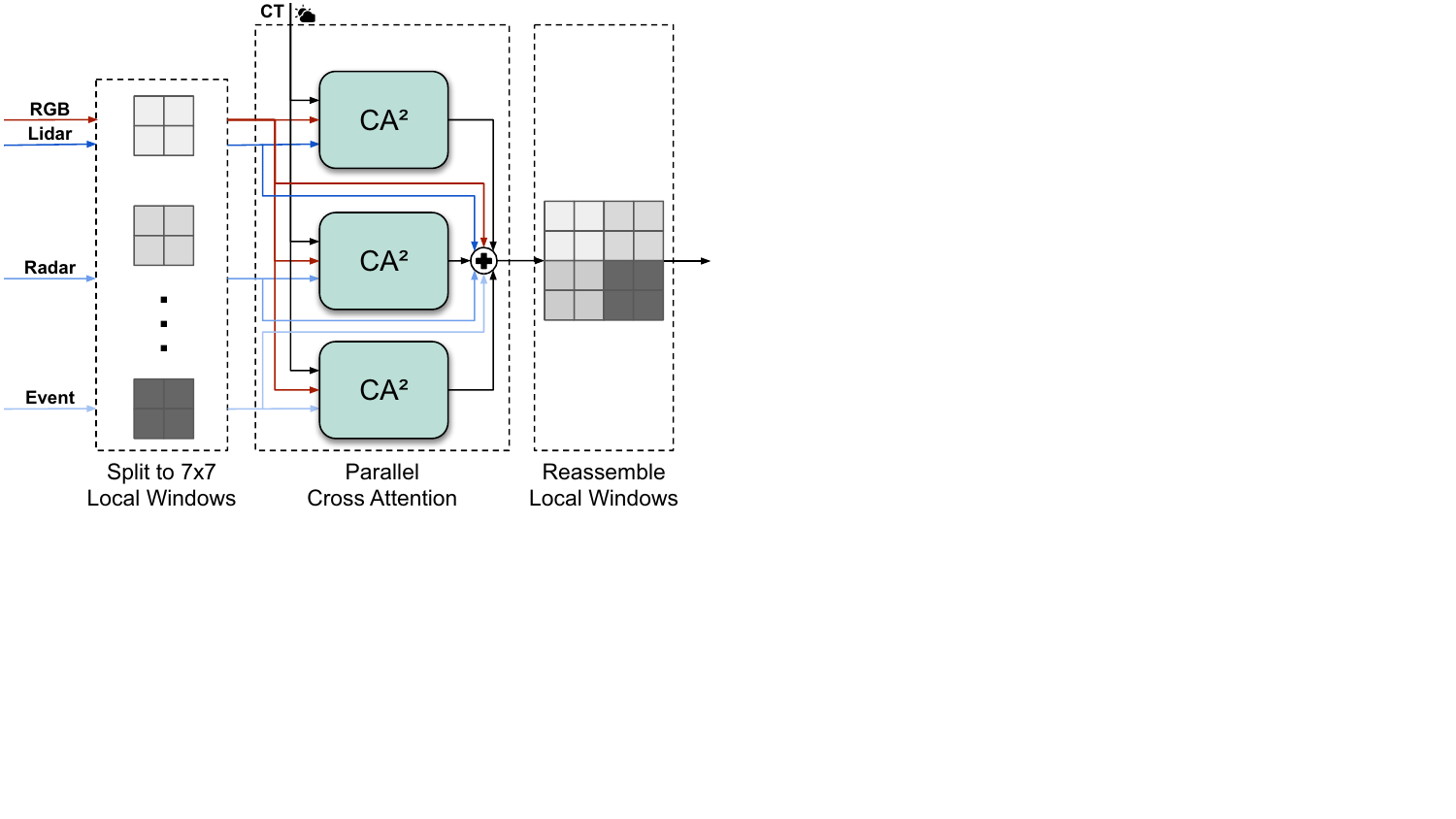}
  \caption{\textbf{Condition-Aware Fusion (CAF)} for our CA$^2$ variant. We apply multi-window cross-attention~\cite{broedermann2023hrfuser} by splitting each modality's feature map into local windows, fusing all secondary modalities in parallel with the RGB features by using our proposed condition-aware cross-attention (CA$^2$) module, and finally stitching the local windows back together.}
  \label{fig:CAF}
\end{figure}

\begin{figure}[tb]
  \centering
  \includegraphics[width=\linewidth,
  trim={0cm 8cm 8cm 0cm}    
  ]{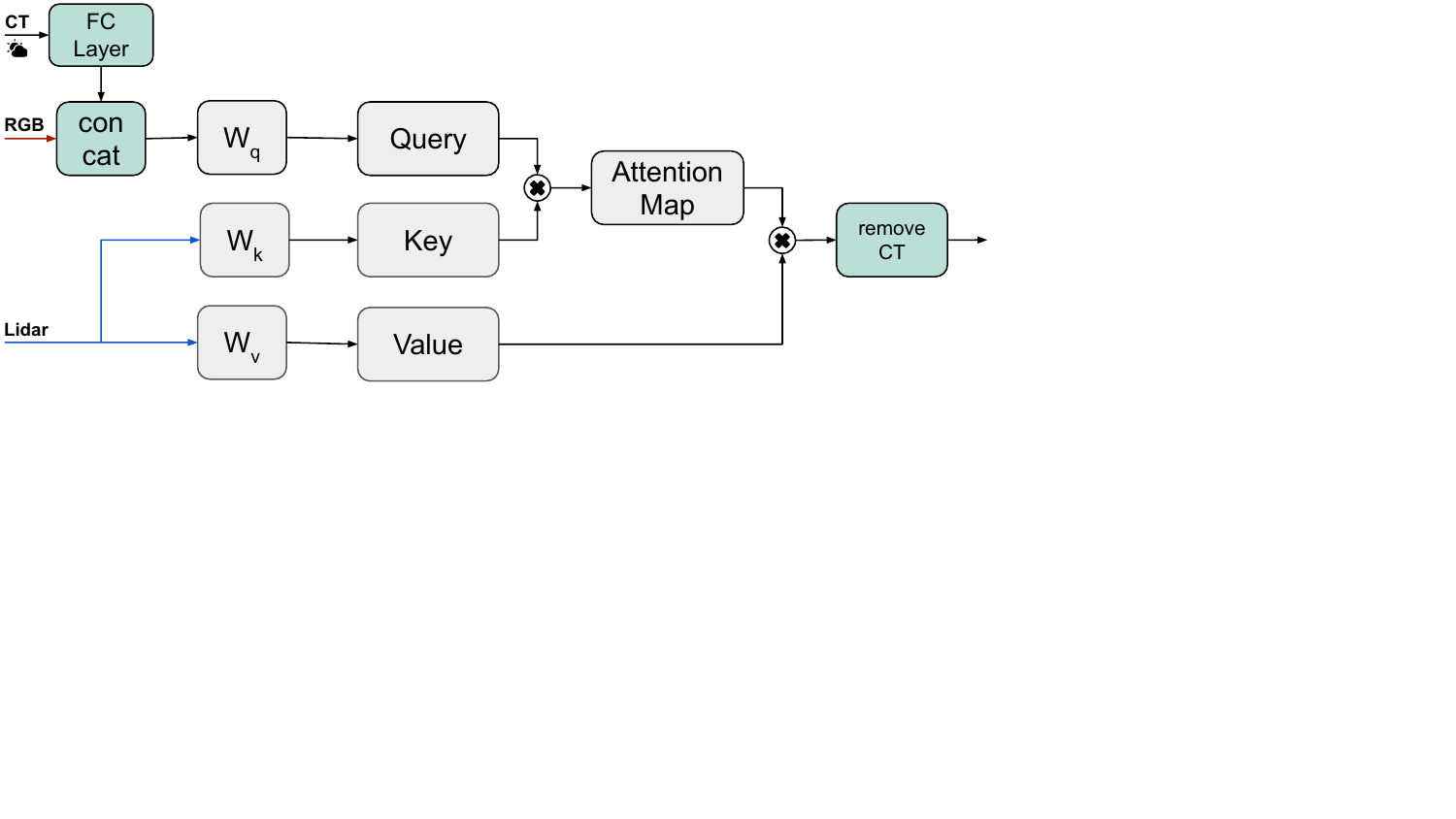}
  \caption{\textbf{Condition-Aware Cross-Attention (CA$^2$)} applied to each local window, here illustrated for the case of RGB-lidar fusion. The Condition Token (CT) is passed through a fully connected layer to align with the feature dimension and concatenated with the RGB tokens to generate a condition-aware query for cross-attention. Afterwards, we remove the token corresponding to the CT to maintain the original spatial dimensions before reassembling
  the full feature map.}
  \label{fig:CA2}
\end{figure}

\PAR{Condition-Aware Addition (CAA) Fusion:} In this simpler approach, we implement a CT-guided weighted addition fusion.
We predict one weight per modality (totaling four modalities), ensuring that all weights sum to one. 
This is achieved by passing the flattened CT through a fully connected layer with four output dimensions, followed by a softmax function.
After the feature adapters, we multiply each modality's feature map by its corresponding predicted weight from the CT. We then sum the weighted feature maps per feature map level.
The fused feature maps are then fed to the OneFormer head.
This method already shows promising results by dynamically increasing the relevance of specific sensors based on different environmental conditions (cf.\ \cref{sec:abl}).

\PAR{Condition-Aware Cross-Attention (CA$^2$) Fusion:}
In our final model, \Ours{}, we introduce CA$^2$ Fusion, where the CT directly guides cross-attention to enhance modality fusion. 
As visualized in \cref{fig:CAF}, we build upon the MWCA fusion block from~\cite{broedermann2023hrfuser}, which utilizes a $7 \times 7$ local-window cross-attention. In this block, we replace the standard cross-attention within each local window with our novel CA$^2$ fusion, integrating the CT into the attention process. 
As depicted in \cref{fig:CA2}, we first pass the CT through a fully connected layer to match the dimensionality of the appropriate feature map tokens. We then concatenate this adjusted CT with the 49 RGB tokens from the local window, forming an enhanced, \emph{condition-aware query} for cross-attention.
This combined query captures both visual features and environmental conditions, which we use to compute the attention map applied to the value of the secondary modality (lidar, radar, etc.).
After obtaining the attention output, we remove the token corresponding to the CT to maintain the original spatial dimensions before reassembling the full feature map.
This condition-aware fusion approach leads to a significant performance boost, as evidenced by our experimental results in \cref{sec:exp}, where we observe notable improvements in segmentation accuracy under challenging conditions.

\PAR{Intuition:}
In our approach, the RGB modality generates a query that interacts with the keys from secondary modalities via cross-attention. The intuition behind this design is that the RGB camera often contains high-quality visual information, but in challenging situations (e.g., a blurry region due to a raindrop on the lens), it might miss critical details. The cross-attention mechanism allows the RGB to ``fill in the blanks'' by looking up at the corresponding regions in the secondary modalities.
Depending on the environmental condition (encoded in the CT) and in contrast to the CAA module, the model can focus more heavily on certain modalities for different regions within an image. For instance, in foggy conditions, the radar might be more reliable for distant objects, while in clear nighttime conditions, the lidar might be more useful. The combination of the RGB and the CT tokens enables the system to generate a dynamic attention map that adapts to the environment, ensuring robust performance across diverse conditions.

\section{Experiments}
\label{sec:exp}

We evaluate \Ours{} on two multimodal driving datasets: MUSES~\cite{MUSES} and DeLiVER~\cite{zhang2023delivering}.
\textbf{MUSES} contains 2500 real-world scenes with panoptic segmentation labels across eight conditions (day and night for rain, snow, fog, and clear weather). Each scene has RGB, lidar, radar, and event camera modalities, along with metadata describing the weather. We use the official SDK to project each modality onto the RGB plane.
\textbf{DeLiVER} offers 7885 synthetic scenes for semantic segmentation in four adverse weather settings (cloudy, foggy, night, rainy), each with five corner-case artifacts (e.g.\ motion blur, lidar jitter). Each scene has RGB, projected lidar, depth, and event camera
modalities. We generate a textual condition prompt similar to Sec.~\ref{sec:methods:caf} with the template: ``A synthetic [\textit{condition}] driving scene with [\textit{case}] artifacts.".

\Ours{} uses a Swin-T backbone~\cite{liu2021swin} and follows the standard OneFormer training setup~\cite{jain2023oneformer} with a batch size of 8.
For MUSES, we train for 960 epochs
with a 20\% random drop of each modality~\cite{MUSES}
and for DeLiVER, we use 200k iterations,
selecting the best checkpoint on the validation set.

\subsection{Comparison to The State of The Art}

We compare our model to state-of-the-art methods both in panoptic and semantic segmentation. Below, we summarize our results in comparison to other methods, showcasing the superior performance of \Ours{}.

In Table~\ref{table:sota:pan:seg}, we compare our model with existing panoptic segmentation methods, both using RGB-camera-only inputs and multimodal inputs. For RGB-camera-only panoptic segmentation, we compare to strong baselines such as Mask2Former~\cite{cheng2022masked}, MaskDINO~\cite{li2023mask}, and OneFormer~\cite{jain2023oneformer}. 
Among these methods, OneFormer achieves the highest performance, highlighting the strength of this architecture.
In a multimodal setting, MUSES~\cite{MUSES} achieves the highest PQ with 53.9\%.
However, our method, which introduces CAF, achieves a \emph{new state-of-the-art} result with a PQ score of 59.7\%.
The performance gain is especially notable in the \textit{night} split with +7.6\% margin over the second best method, compared to +1.9\% in the \textit{day} split. Since each time-of-day split includes all weather conditions (clear, fog, rain, and snow), these findings underscore the importance of adapting sensor weights when the RGB modality is less reliable.

As multimodal panoptic segmentation is still an emerging field, we further benchmark our model against the state of the art in multimodal \emph{semantic} segmentation on both MUSES in Table~\ref{table:sota:sem:seg:muses} and DeLiVER in Table~\ref{table:sota:sem:seg:deliver}. Given the OneFormer head simultaneously solves panoptic and semantic segmentation, we obtain one \Ours{} model solving both tasks on MUSES. The results show that our model outperforms top methods such as CMNeXt~\cite{zhang2023delivering} and GeminiFusion~\cite{jia2024geminifusion}. While these multimodal methods do improve over the RGB-camera-only methods, our \Ours{} model significantly outperforms all multimodal and RGB-only methods and sets the new SOTA.
This validates the strength of CAF, which allows optimal sensor integration under diverse environmental conditions and thus superior performance both in panoptic and semantic segmentation.

\begin{table}
  \caption
  {Comparison of panoptic segmentation methods on the MUSES test set in PQ$\uparrow$. 
  $^{*}$: uses only the camera modality as input.
  }
  \label{table:sota:pan:seg}
  \smallskip
  \centering
  \setlength\tabcolsep{4pt}
  \scriptsize
  \begin{tabular*}{\linewidth}{@{\extracolsep{\fill}} lcccccccc@{}}
  \toprule
  \textbf{Method}&  \textbf{Clear} & \textbf{Fog} & \textbf{Rain} & \textbf{Snow} & \textbf{Day} & \textbf{Night} & \textbf{All} \\
   \hline
   Mask2Former~\cite{cheng2022masked}$^{*}$  &48.8  &46.5  &45.4  &45.1  &49.4&39.4&46.9 \\
   MaskDINO~\cite{li2023mask}$^{*}$  &54.1&46.2&46.23&48.54&51.9&42.7& 49.4 \\ 
   OneFormer~\cite{jain2023oneformer}$^{*}$     &58.3  &53.7  &53.4  &53.8  &57.6&47.8&55.2  \\
   HRFuser~\cite{broedermann2023hrfuser}    &47.0  &43.6  &42.7  &40.6  &44.6&40.0&43.9 \\ 
   MUSES~\cite{MUSES}    &55.3  &50.3  &53.8  &50.5  &54.1&49.7&53.6 \\
   \hline
   \Ours{}-CAA  &\underline{61.2}&\underline{56.4}&\underline{59.4}&\textbf{57.9}&\textbf{59.9}&\underline{56.2}& \underline{59.4} \\ 
    \Ours{}-CA$^2$~(\textbf{Ours})    &\textbf{61.4}   &\textbf{57.5}   &\textbf{59.6}   &\underline{57.2} &\underline{59.5}&\textbf{57.3}&\textbf{59.7} \\ 
  \bottomrule  
  \end{tabular*}
\end{table}

\begin{table}
  \caption
  {Comparison of semantic segmentation methods on the MUSES test set. C: RGB Camera, L: Lidar, R: Radar, E: Events}
  \label{table:sota:sem:seg:muses}
  \smallskip
  \centering
  \setlength\tabcolsep{4pt}
  \scriptsize
  \begin{tabular*}{\linewidth}{l @{\extracolsep{\fill}} ccc}
  \toprule
  \textbf{Method} & \textbf{Modalities} & \textbf{Backbone} & \textbf{mIoU} $\uparrow$ \\
   \hline
   Mask2Former~\cite{cheng2022masked} & C & Swin-T & 70.7  \\ 
   SegFormer~\cite{xie2021segformer} & C & MiT-B2 &  72.5  \\ 
   OneFormer~\cite{jain2023oneformer} & C & Swin-T  &  72.8  \\ 
   CMNeXt~\cite{zhang2023delivering} & CLRE & MiT-B2 &  72.4\\ 
   GeminiFusion~\cite{jia2024geminifusion} & CLRE & MiT-B2 & 75.3 \\ 
   \hline
   \Ours{}-CAA & CLRE & Swin-T & \textbf{78.5} \\
   \Ours{}-CA$^2$~(\textbf{Ours}) & CLRE & Swin-T & \underline{78.2} \\
  
  \bottomrule  
  \end{tabular*}
\end{table}

\begin{table}
  \caption
  {Comparison of semantic segmentation methods on DeLiVER. C: RGB Camera, D: Depth, E: Events, L: Lidar}
  \label{table:sota:sem:seg:deliver}
  \smallskip
  \centering
  \setlength\tabcolsep{4pt}
  \scriptsize
  \begin{tabular*}{\linewidth}{l @{\extracolsep{\fill}} cccc}
  \toprule
  \textbf{Method} & \textbf{Modalities} & \textbf{Backbone} & \textbf{mIoU-val} $\uparrow$ & \textbf{mIoU-test} $\uparrow$\\
   \hline
   CMNeXt~\cite{zhang2023delivering} & CLDE & MiT-B2 &  66.3 & 53.0\\ 
   StitchFusion~\cite{li2024stitchfusion} & CLDE & MiT-B2 & 68.2 & 53.4 \\ 
   GeminiFusion~\cite{jia2024geminifusion} & CLDE & MiT-B2 & 66.9 & 54.5\\ 
   \hline
   \Ours{}-CAA& CLDE & Swin-T & 68.6  & \underline{55.2}\\ 
   \Ours{}-CA$^2$~(\textbf{Ours}) & CLDE & Swin-T & 67.8  & \textbf{55.6}\\   
  \bottomrule  
  \end{tabular*}
\end{table}

\subsection{Ablation Studies}
\label{sec:abl}

We perform all our ablations on MUSES due to its diverse real-world conditions and high-quality panoptic annotations.

\PAR{Feature Adapter:}
We ablate the effect of our proposed feature adapter and CAF module in~\cref{table:abl:modules} where we first create a strong baseline (row 2) by training OneFormer with the 4 parallel backbones proposed by MUSES. Using a shared backbone (row 3) significantly reduces the number of parameters (-55\%), but also results in a significant performance drop (-0.9\% PQ). Adding our proposed feature adapter (row 4) gains back all the lost performance (+0.9\%  PQ) while still having less than half of the original parameters. Adding our CAF module (row 5) gains another 0.4\% in PQ, surpassing the performance of the larger, baseline model.

\begin{table}
  \caption
  {Ablation study of our proposed modules on MUSES. Using a shared backbone reduces the parameters by 55\% while the adapter and CAF module increase the PQ to a new SOTA.
  }
  \label{table:abl:modules}
  \smallskip
  \centering
  \setlength\tabcolsep{4pt}
  \scriptsize
  \begin{tabular*}{\linewidth}{l @{\extracolsep{\fill}} lccccccc}
  \toprule
   & \textbf{Method} & \textbf{Shared Backb.} & \textbf{Adapter} & \textbf{CAF} & \textbf{Params} & \textbf{PQ} $\uparrow$ \\
   \hline
    1~ & OneFormer~\cite{jain2023oneformer} & n/a &  - & - &  50.7M  &55.7   \\     
    2 & OneFormer w/ MUSES  & - &  - & - & 149.0M  & \underline{59.3}\label{table:abl:modules:OFM} \\ 
    3  & \Ours{} & \checkmark & - & - & 66.4M  & 58.4 \\    
    4 & \Ours{} & \checkmark& \checkmark & - & 68.0M & \underline{59.3}\\
    5 & \Ours{}-CA$^2$~(\textbf{Ours}) & \checkmark & \checkmark & \checkmark & 77.7M & \textbf{59.7}\\
  \bottomrule  
  \end{tabular*}
\end{table}

\PAR{CAA Fusion:} 
We conduct an ablation study to evaluate the impact of our CAA fusion method. As shown in Table~\ref{tab:abl:caa}, we compare four fusion strategies: mean, random weights, learned weights, and CAA Fusion, which dynamically predicts modality weights using the CT. 
For the basic method that takes a simple mean over all modalities, we can see a large improvement of +1\% PQ when adding the adapter (row 2) to the no-adapter baseline, highlighting the benefit of adapting features before fusion. Random fusion, where weights are randomly assigned to each modality, underperformed with a PQ of 58.5\%. Learning static weights for each modality, without considering environmental conditions, reaches a similar performance to the mean fusion. Finally, our proposed CAA Fusion, which dynamically adjusts weights based on the CT, outperforms all other strategies with a PQ of 59.4\%, demonstrating the effectiveness of adapting the fusion process to environmental conditions.

\begin{table}[]
\centering
\caption{Ablation study using all four modalities on different weighted fusion strategies for our CAA module on MUSES.}
\label{tab:abl:caa}
\begin{tabular}{lcccc}
\toprule
& \textbf{Adapter}      & \textbf{CAF} & \textbf{Fusion Type} & \textbf{PQ} $\uparrow$ \\
\hline 
1          & \no                                          & \no               & Mean             & 58.1  \\ 
2           & \yes                                          & \no               & Mean          & 59.1 \\
3           & \yes                                          & \no               & Random Weights           & 58.5\\
4           & \yes                                          & \no               & Learned Weights                &        59.1     \\
5           & \yes                                          & \yes               & CAA               & \textbf{59.4} \\

\bottomrule
\end{tabular}
\end{table}

\PAR{CA$^2$ Fusion:}
In \cref{tab:abl:ca2}, we investigate the detailed design of our CA$^2$ fusion and how to best utilize the CT for CAF. 
Appending the CT to the secondary modalities' keys and values yields a small decrease in performance.
Since the CT is derived from RGB features, mixing it with other modalities could introduce confusion to the fusion process.

In contrast, our proposed approach (row 3), where the CT is only appended to the RGB queries, achieves the highest PQ score of 59.7\%. 
This design enables dynamic cross-attention tailored to environmental conditions, effectively guiding the fusion process and aligning with our intuition described in \cref{sec:methods:caf} and motivating our design choice.

\begin{table}[]
\centering
\caption{Ablation study on CA$^2$ fusion design on MUSES. The Condition Token is appended to the RGB queries (Q) or the secondary modalities' keys and values (K\&V).}
\label{tab:abl:ca2}
\begin{tabular}{lccc}
\toprule
 & \textbf{Q}                  & \textbf{K\&V}        & \textbf{PQ} $\uparrow$ \\
\hline
1 & \no                  & \no             &   59.3 \\
2 &  \no                  & \yes             &  59.1  \\
 3 (\textbf{Ours})  & \yes                  & \no             &  \textbf{59.7}  \\
 4   & \yes                  & \yes             &  59.6  \\
\bottomrule
\end{tabular}
\end{table}

\PAR{Condition Loss:}
In Table~\ref{tab:abl:loss}, we assess the impact of our contrastive condition loss, by training an identical network, with and without applying the loss on the CT. As this loss is designed to ensure the CT effectively captures environmental conditions, we want to investigate if the model could learn this internally without additional supervision. Adding the condition loss improves the performance by +0.4\% in PQ. This demonstrates that our contrastive loss is essential as an additional supervision to efficiently guide CAF.

\begin{table}[]
\centering
\caption{Ablation study on the guidance of the verbo-visual contrastive loss between $Q_{text}$ and the CT on MUSES.}
\label{tab:abl:loss}
\begin{tabular}{lcc}
\toprule
 & \textbf{Condition Loss}                            & \textbf{PQ} $\uparrow$ \\
\hline
 1 & \no &   59.3 \\
  2 (\textbf{Ours}) &   \yes &  \textbf{59.7} \\
\bottomrule
\end{tabular}
\end{table}

\begin{figure}[tb]
  \centering
  \resizebox{\linewidth}{!}{  
  \begin{tikzpicture}[baseline]
    \begin{axis}[
      xbar stacked,
      width=\linewidth,
      height=5cm,
      bar width=0.3cm,
      axis line style={draw=none},
      symbolic y coords={Fog Night, Rain Night, Snow Night, Clear Night, Fog Day, Rain Day, Snow Day, Clear Day},
      ytick=data,
      enlarge y limits=0.12,
      xmin=0,
      xmax=100,
      font=\footnotesize,
      legend style={at={(0.5,-0.1)}, anchor=north, draw=none},
      legend cell align={left},
      legend columns=-1,
      nodes near coords,
      every node near coord/.append style={font=\scriptsize,white},
      ]
            
      \addplot+[xbar,fill=color1!80,draw=black] 
      plot coordinates {
        (68,Clear Day)
        (69,Fog Day)
        (63,Rain Day)
        (67,Snow Day)
        (57,Clear Night)
        (48,Fog Night)
        (54,Rain Night)
        (60,Snow Night)
      };

      \addplot+[xbar,fill=color2!80,draw=black] 
      plot coordinates {
        (19,Clear Day)
        (18,Fog Day)
        (19,Rain Day)
        (18,Snow Day)
        (23,Clear Night)
        (27,Fog Night)
        (23,Rain Night)
        (22,Snow Night)
      };

      \addplot+[xbar,fill=color3!80,draw=black] 
      plot coordinates {
        (5,Clear Day)
        (5,Fog Day)
        (6,Rain Day)
        (6,Snow Day)
        (7,Clear Night)
        (7,Fog Night)
        (7,Rain Night)
        (6,Snow Night)
      };
      
      \addplot+[xbar,fill=color4!80,draw=black] 
      plot coordinates {
        (8,Clear Day)
        (8,Fog Day)
        (12,Rain Day)
        (9,Snow Day)
        (13,Clear Night)
        (18,Fog Night)
        (16,Rain Night)
        (12,Snow Night)
      };
            
      \legend{RGB, Lidar, Radar, Events}
    \end{axis}
  \end{tikzpicture}
  }
  \caption{
  Average Condition-Aware Addition (CAA) fusion weights in \% on the MUSES test set across different weather conditions and times of day. This figure illustrates how the relative contributions of each sensor modality vary under various environmental conditions, highlighting the adaptability of the fusion mechanism to changing visibility and lighting scenarios.}
  \label{fig:abl:caa:weights:all}
\end{figure}
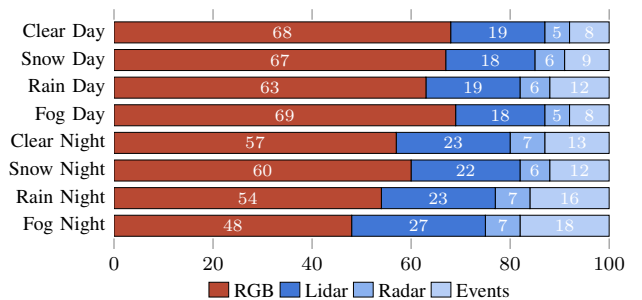

\begin{figure*}[btp]
\centering
\begin{tabular}{@{}c@{\hspace{0.03cm}}c@{\hspace{0.03cm}}c@{\hspace{0.03cm}}c@{\hspace{0.03cm}}c@{\hspace{0.03cm}}c@{\hspace{0.03cm}}c@{\hspace{0.03cm}}c@{}}
\subfloat{\scriptsize RGB} &
\subfloat{\scriptsize Lidar} &
\subfloat{\scriptsize Radar} &
\subfloat{\scriptsize Events} &
\subfloat{\scriptsize Ground Truth} &
\subfloat{\scriptsize OneFormer~\cite{jain2023oneformer}} &
\subfloat{\scriptsize OneFormer-MUSES} &
\subfloat{\scriptsize \Ours{}} \\
\vspace{-0.1cm}

\includegraphics[width=0.12\textwidth]{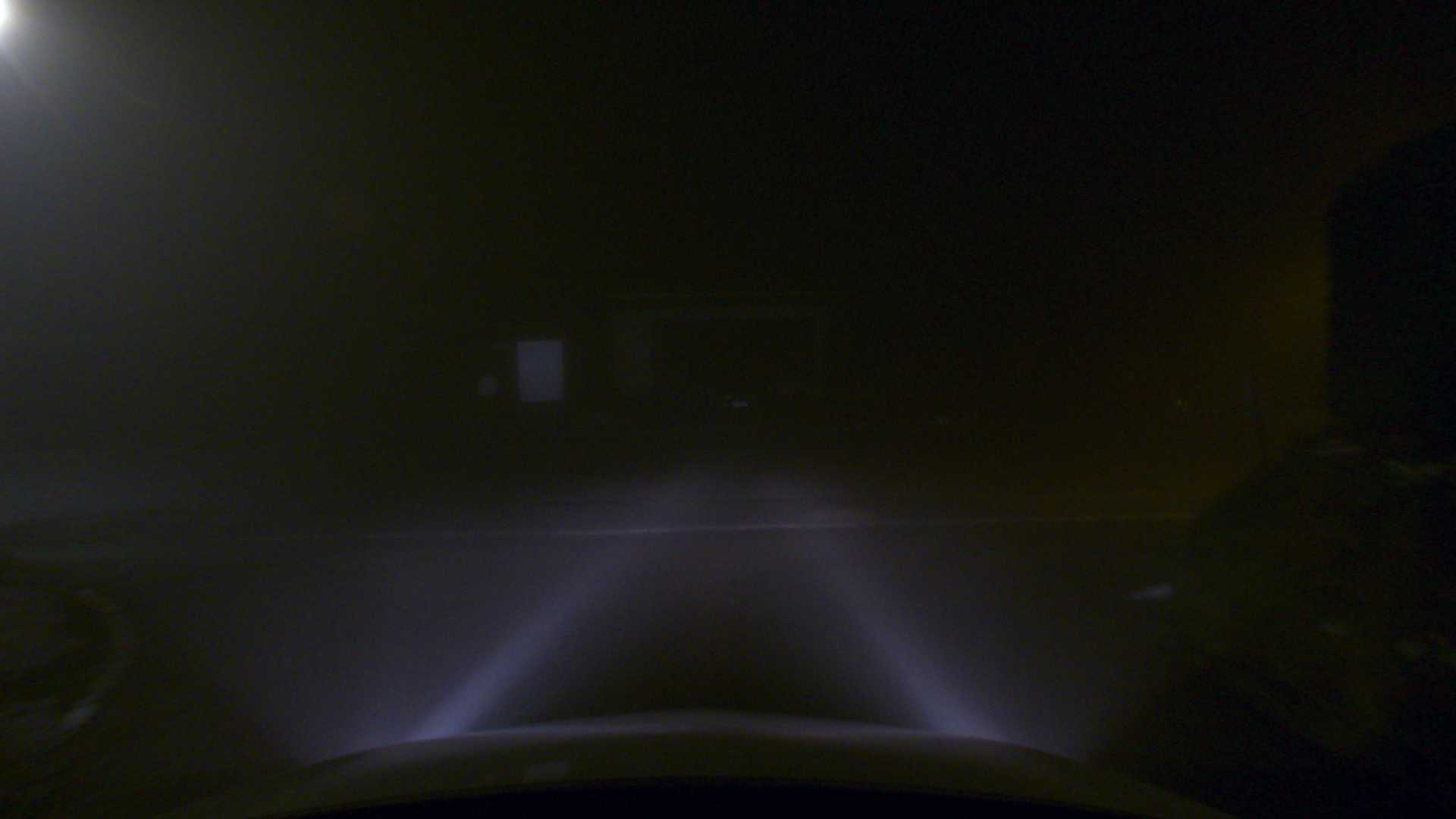} &
\includegraphics[width=0.12\textwidth]{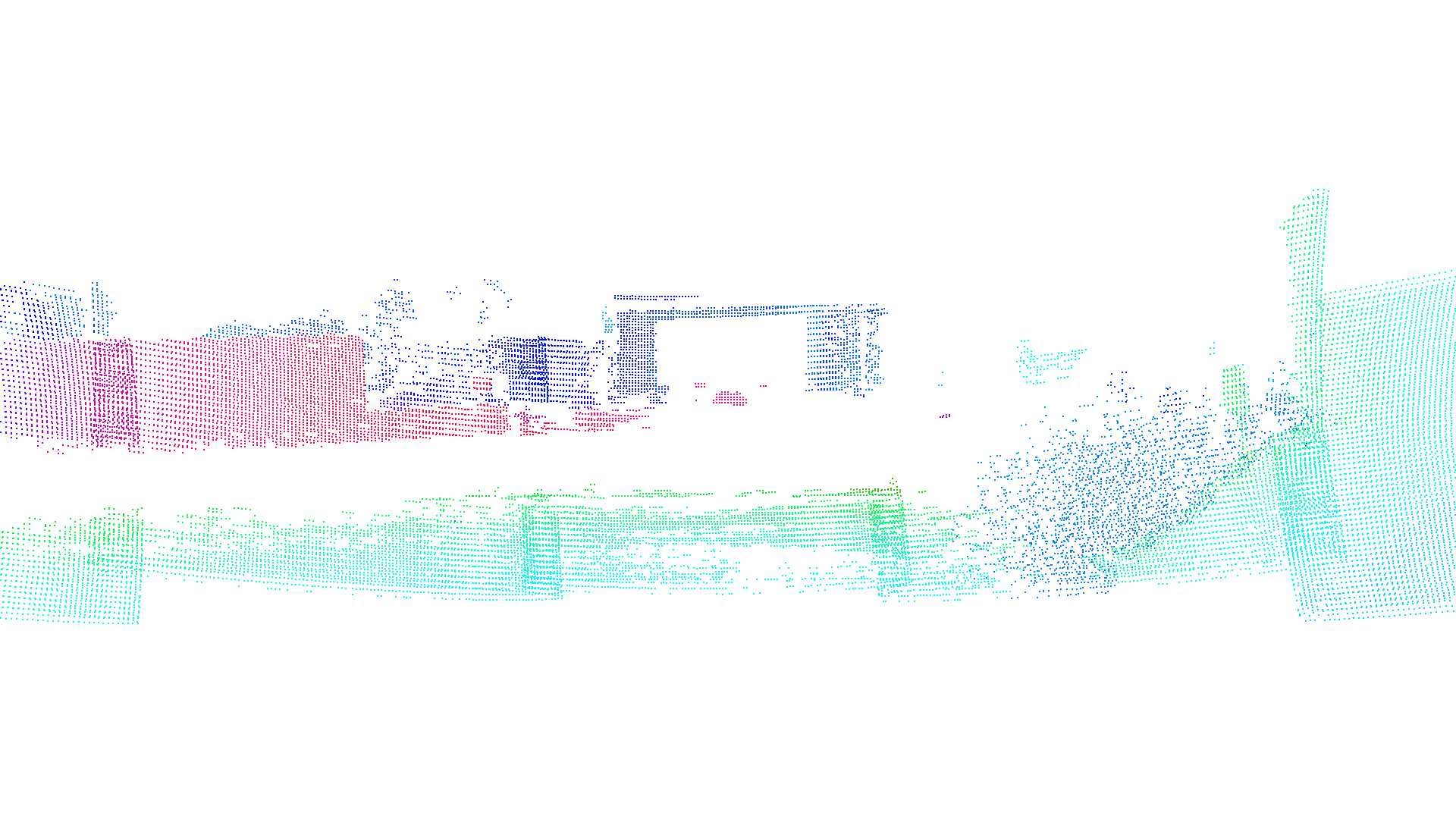} &
\includegraphics[width=0.12\textwidth]{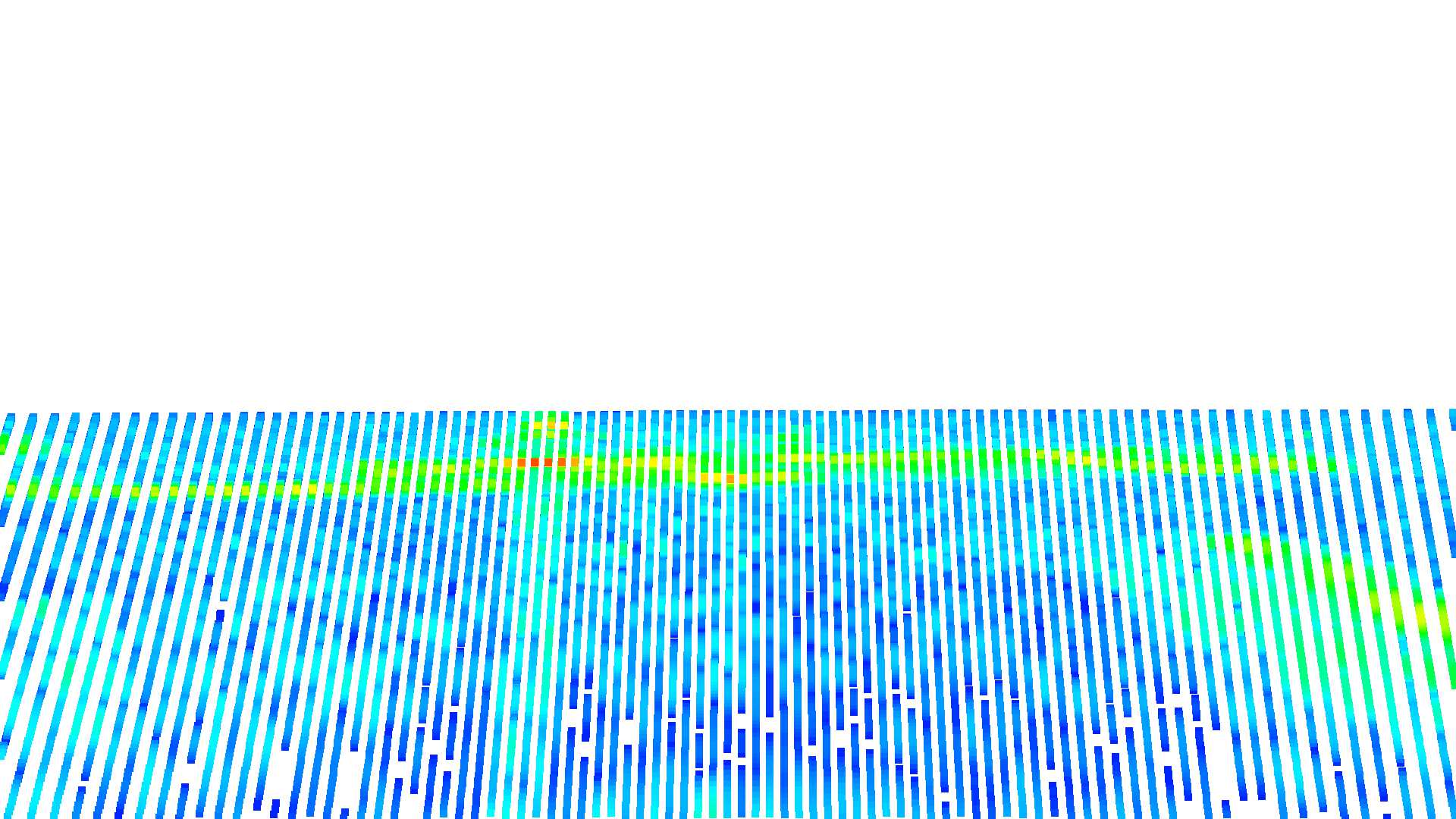} &
\includegraphics[width=0.12\textwidth]{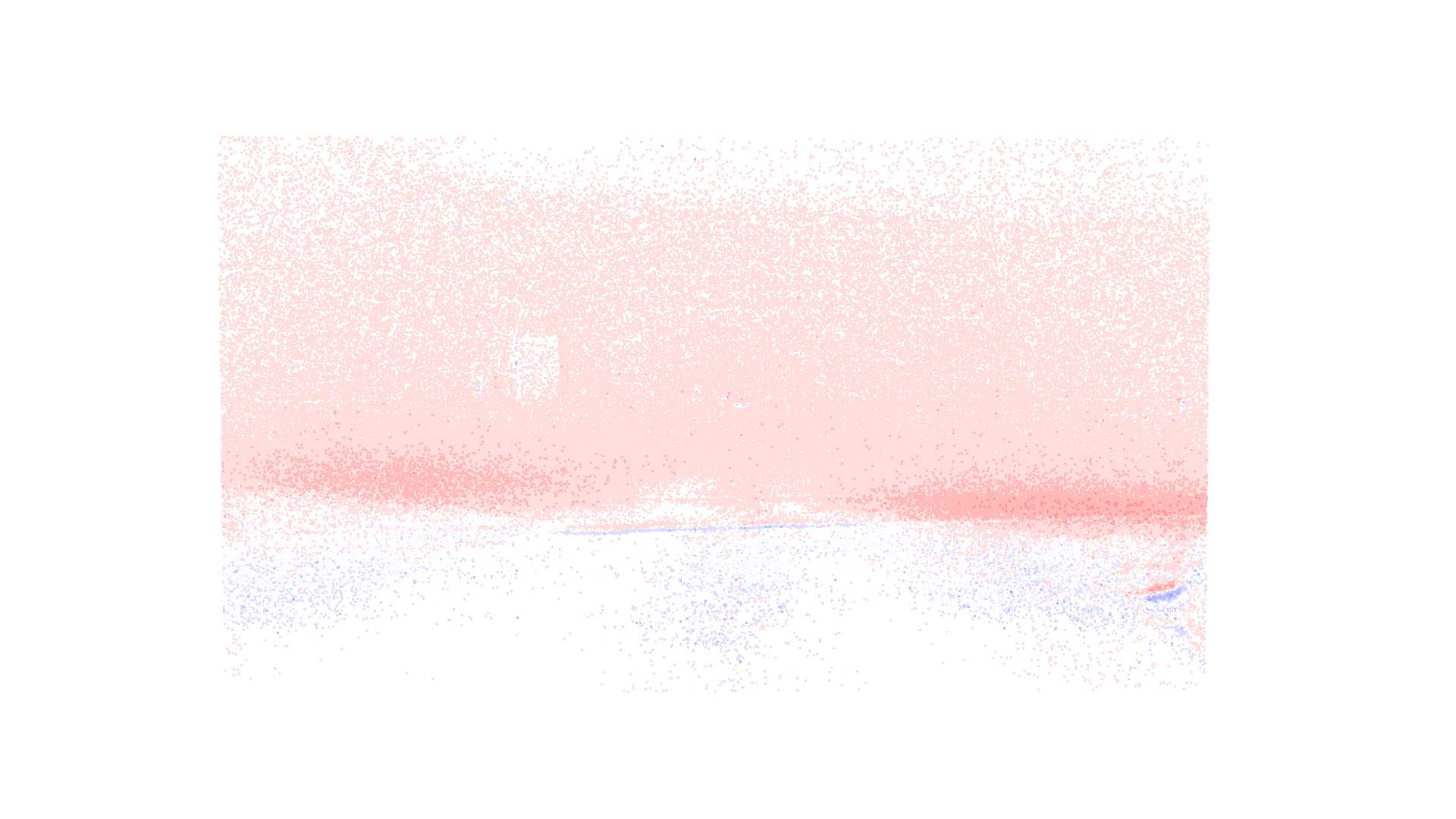} &
\includegraphics[width=0.12\textwidth]{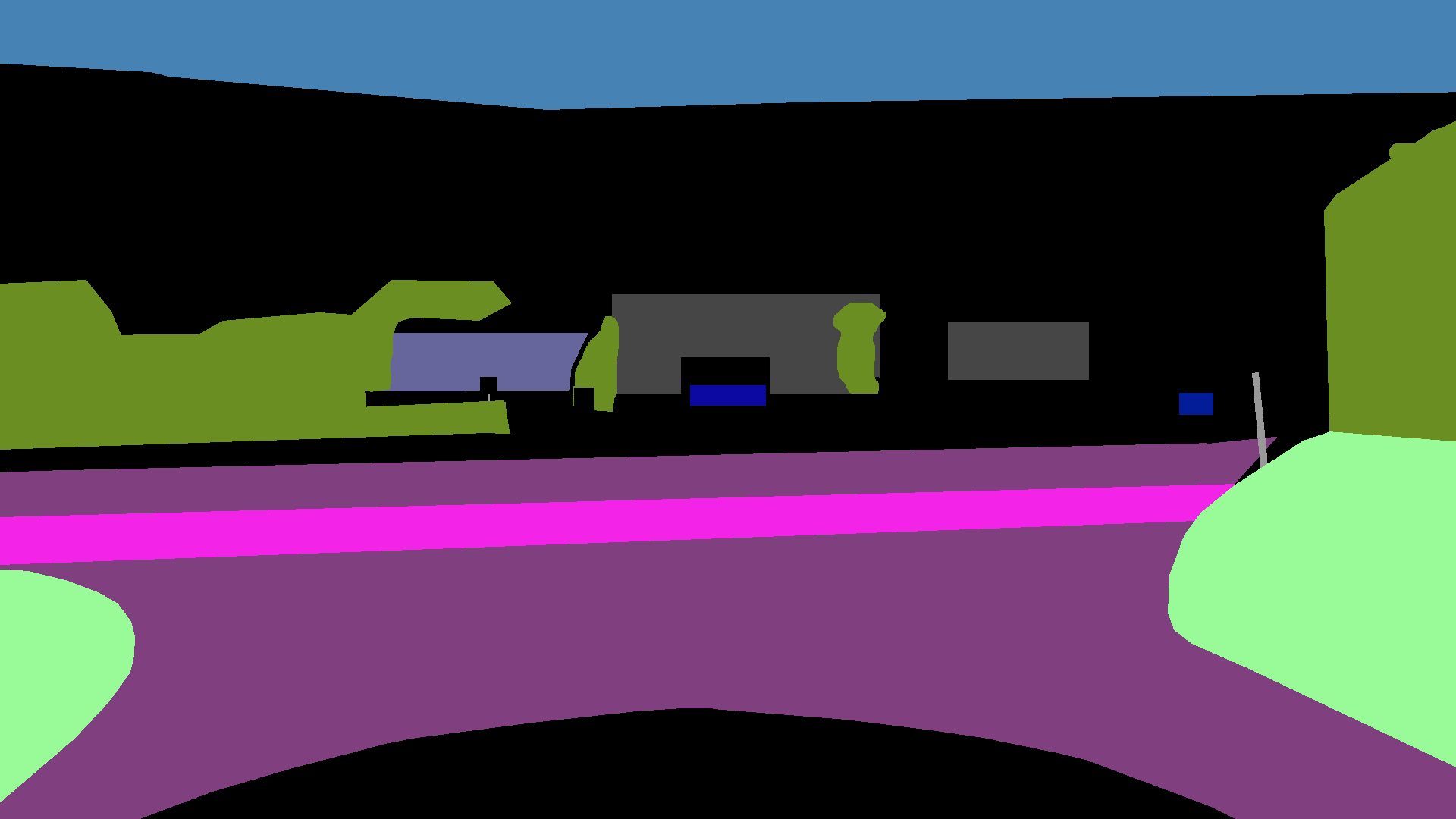} &
\includegraphics[width=0.12\textwidth]{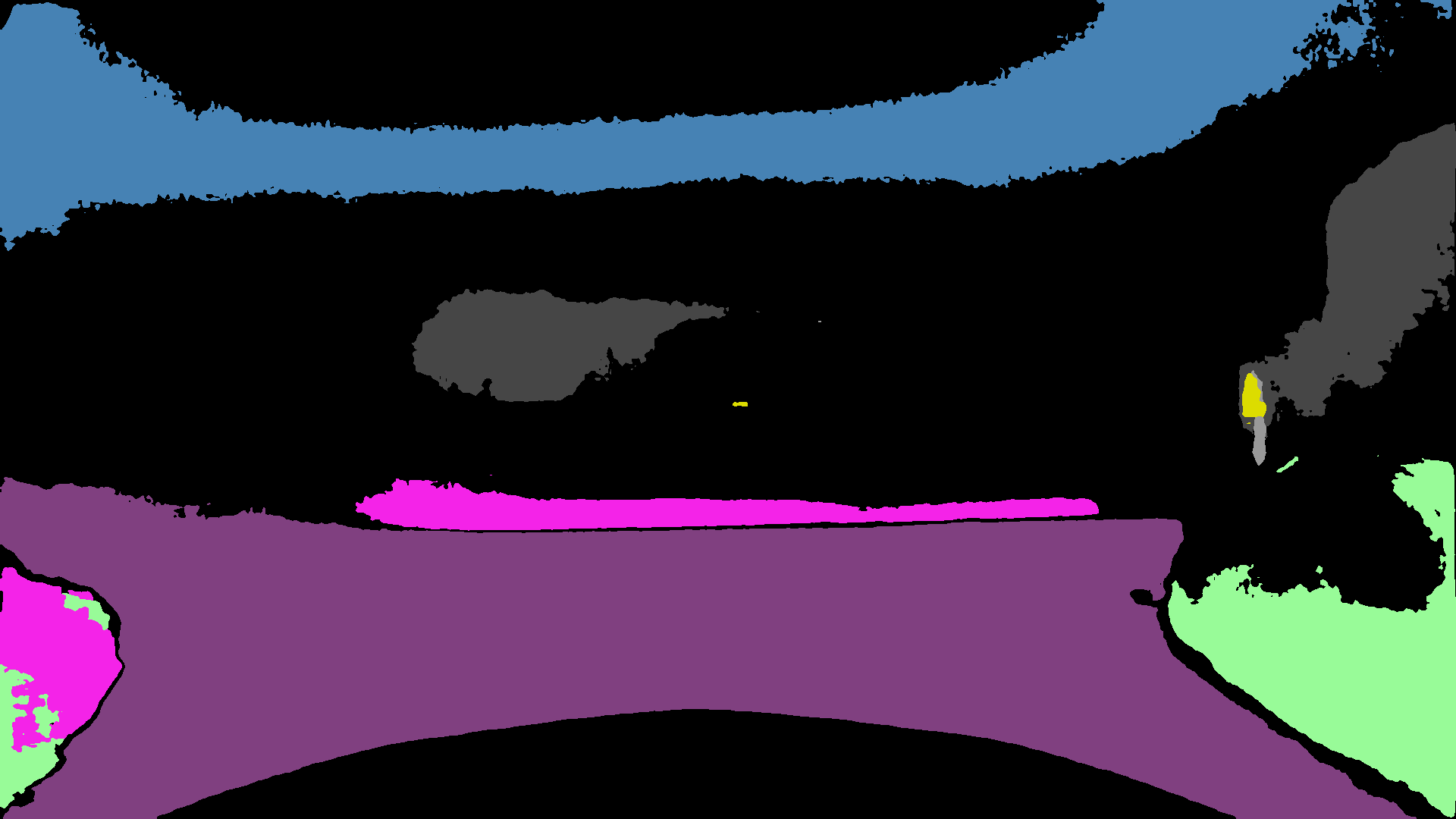} &
\includegraphics[width=0.12\textwidth]{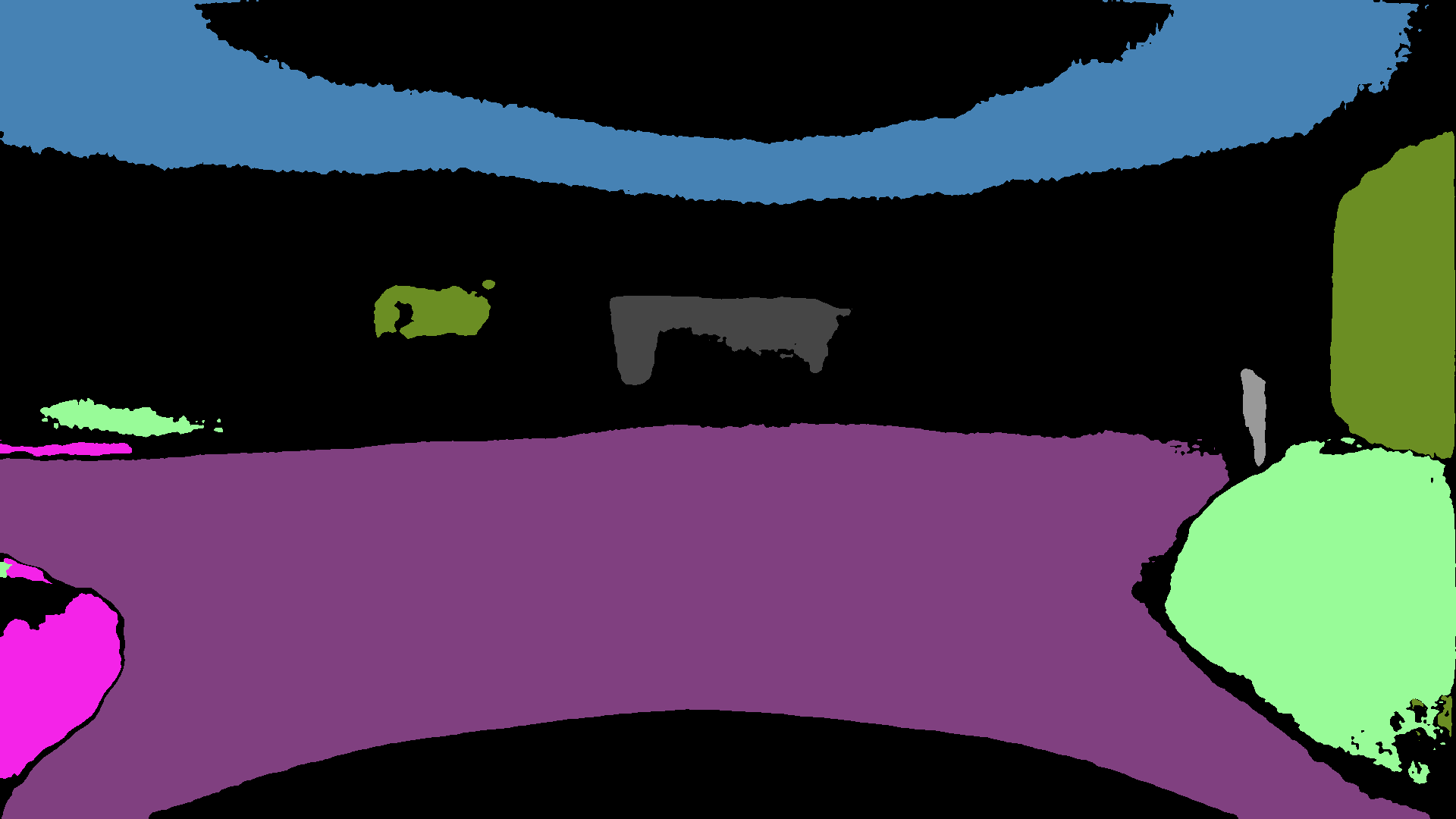} &
\includegraphics[width=0.12\textwidth]{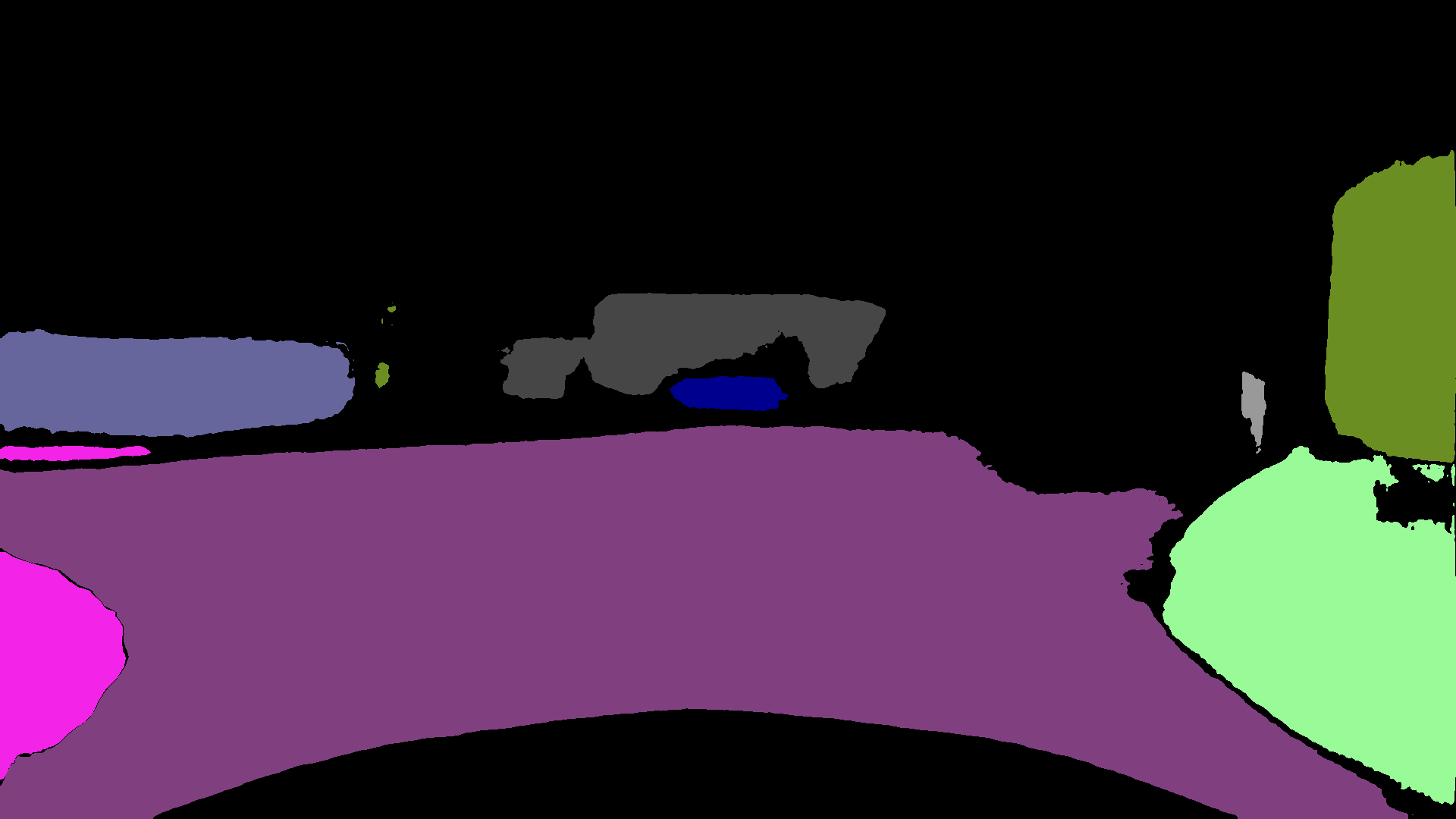} \\
\vspace{-0.1cm}

\includegraphics[width=0.12\textwidth]{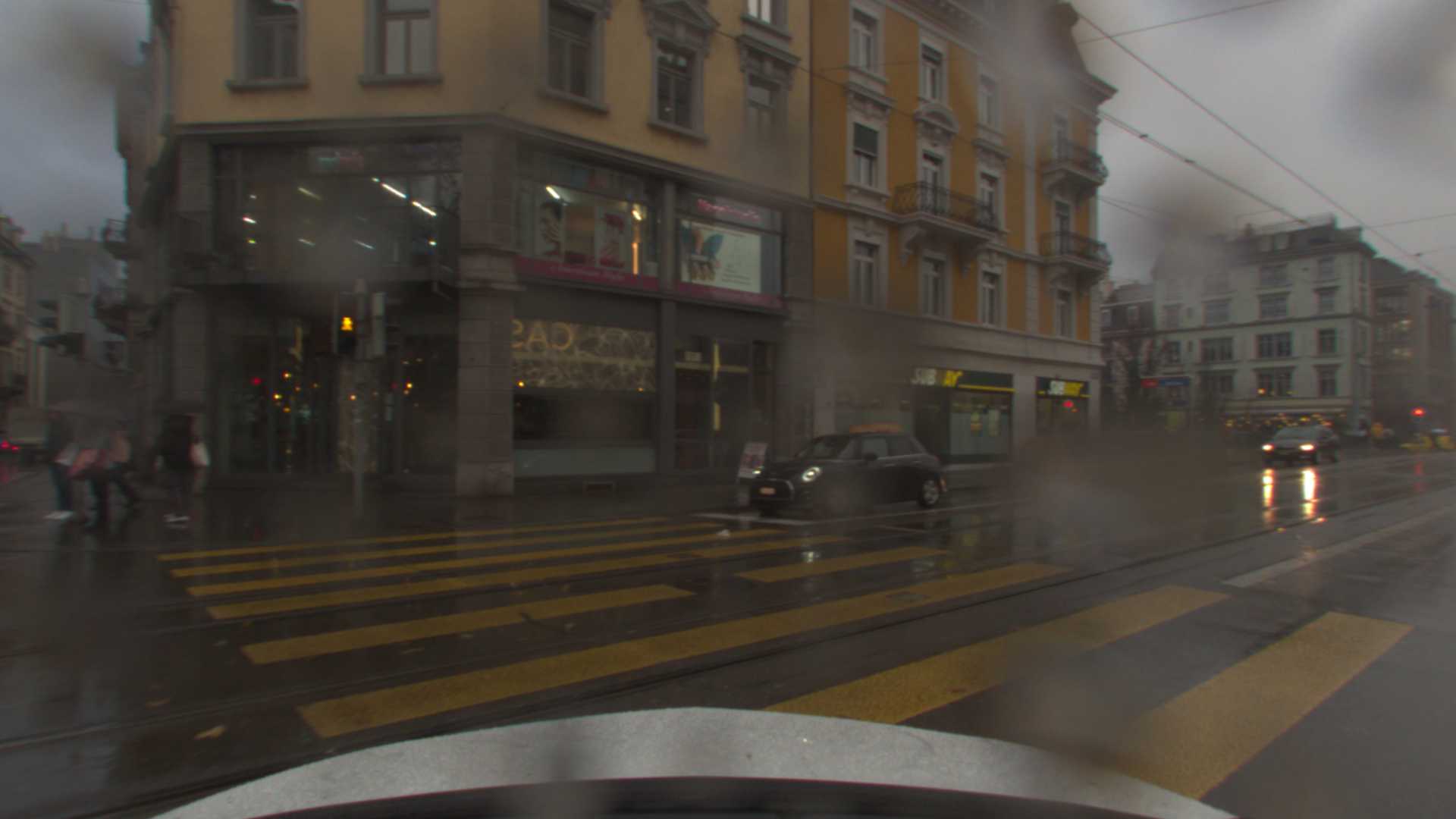} &
\includegraphics[width=0.12\textwidth]{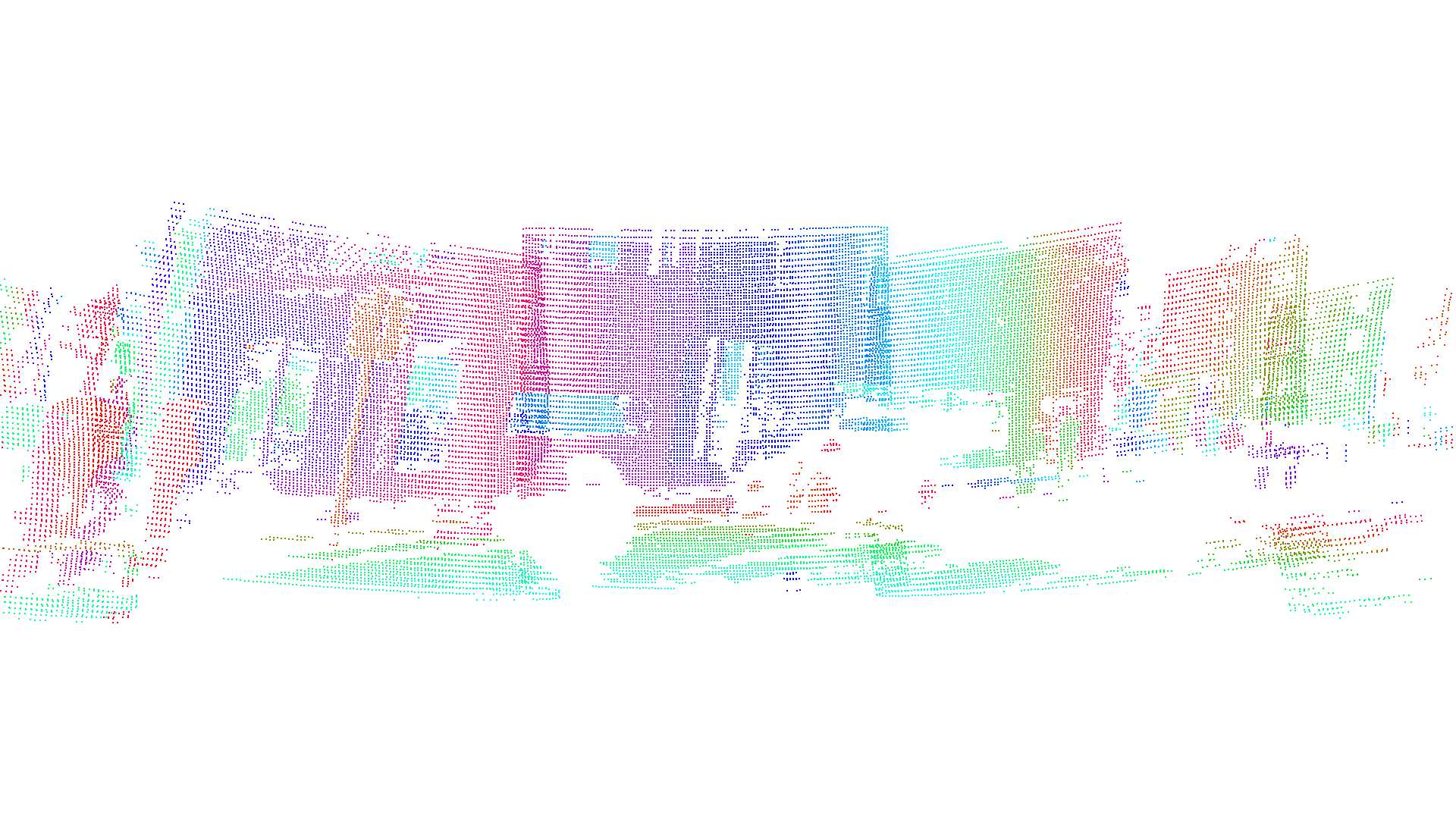} &
\includegraphics[width=0.12\textwidth]{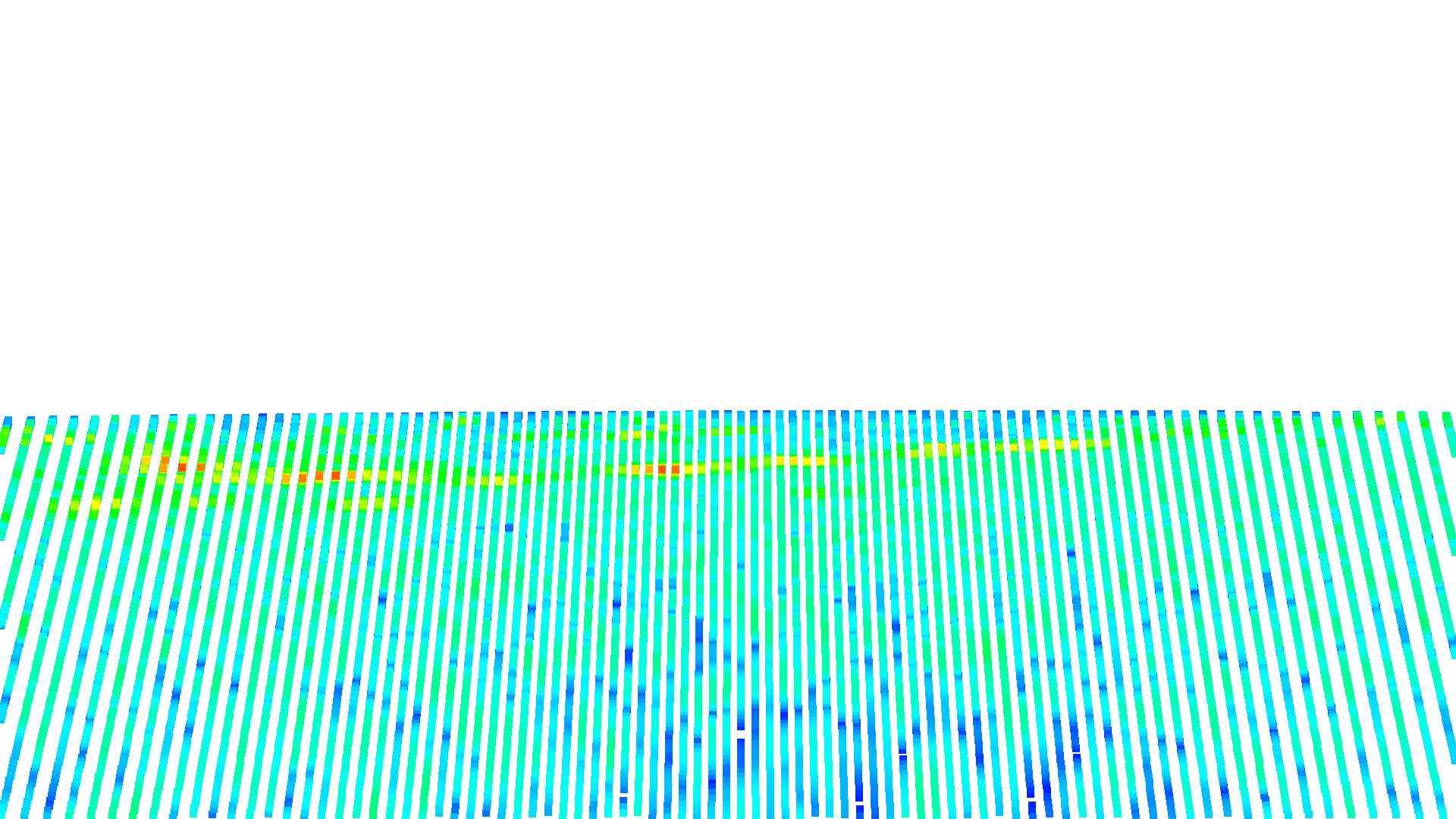} &
\includegraphics[width=0.12\textwidth]{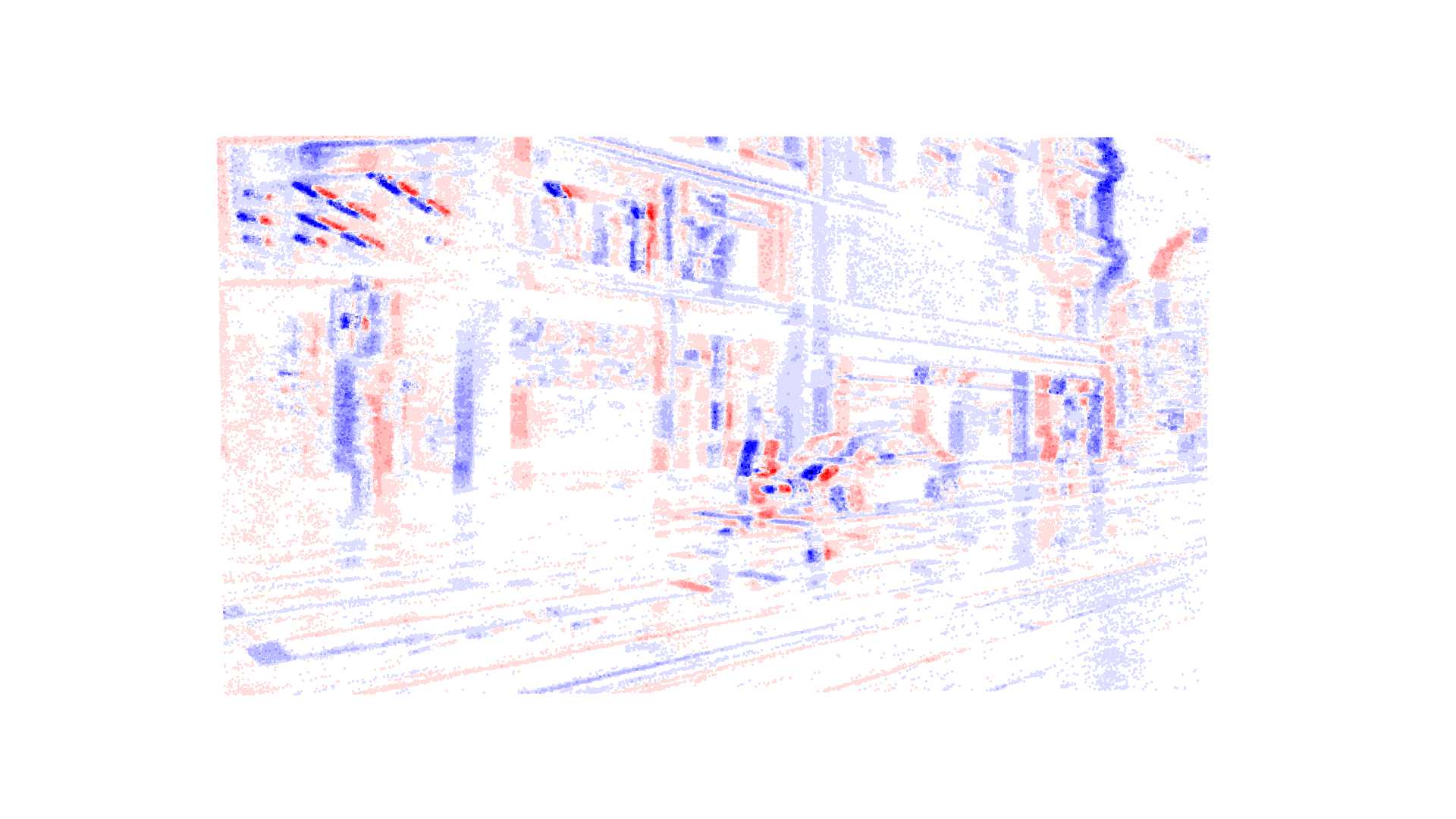} &
\includegraphics[width=0.12\textwidth]{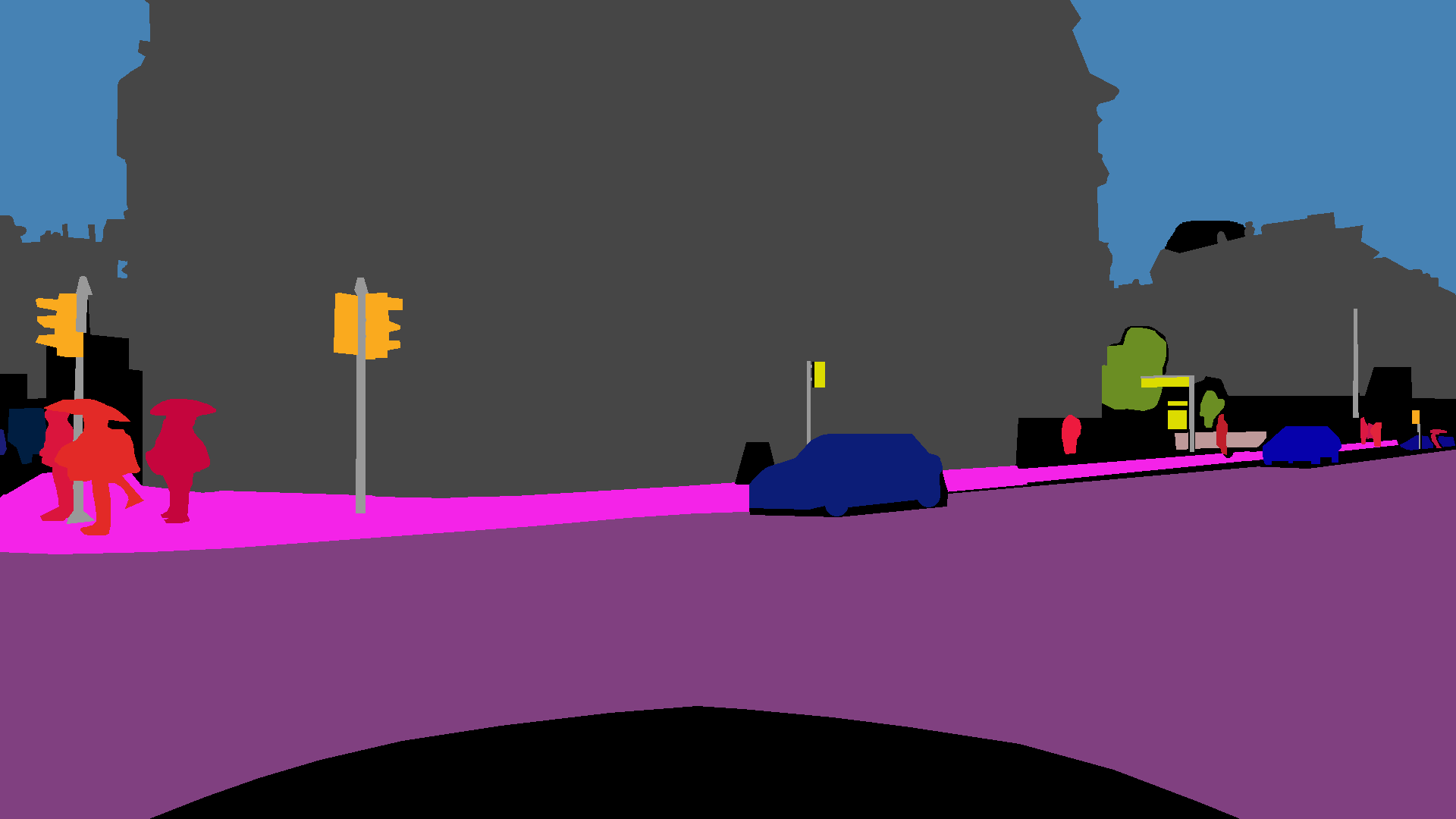} &
\includegraphics[width=0.12\textwidth]{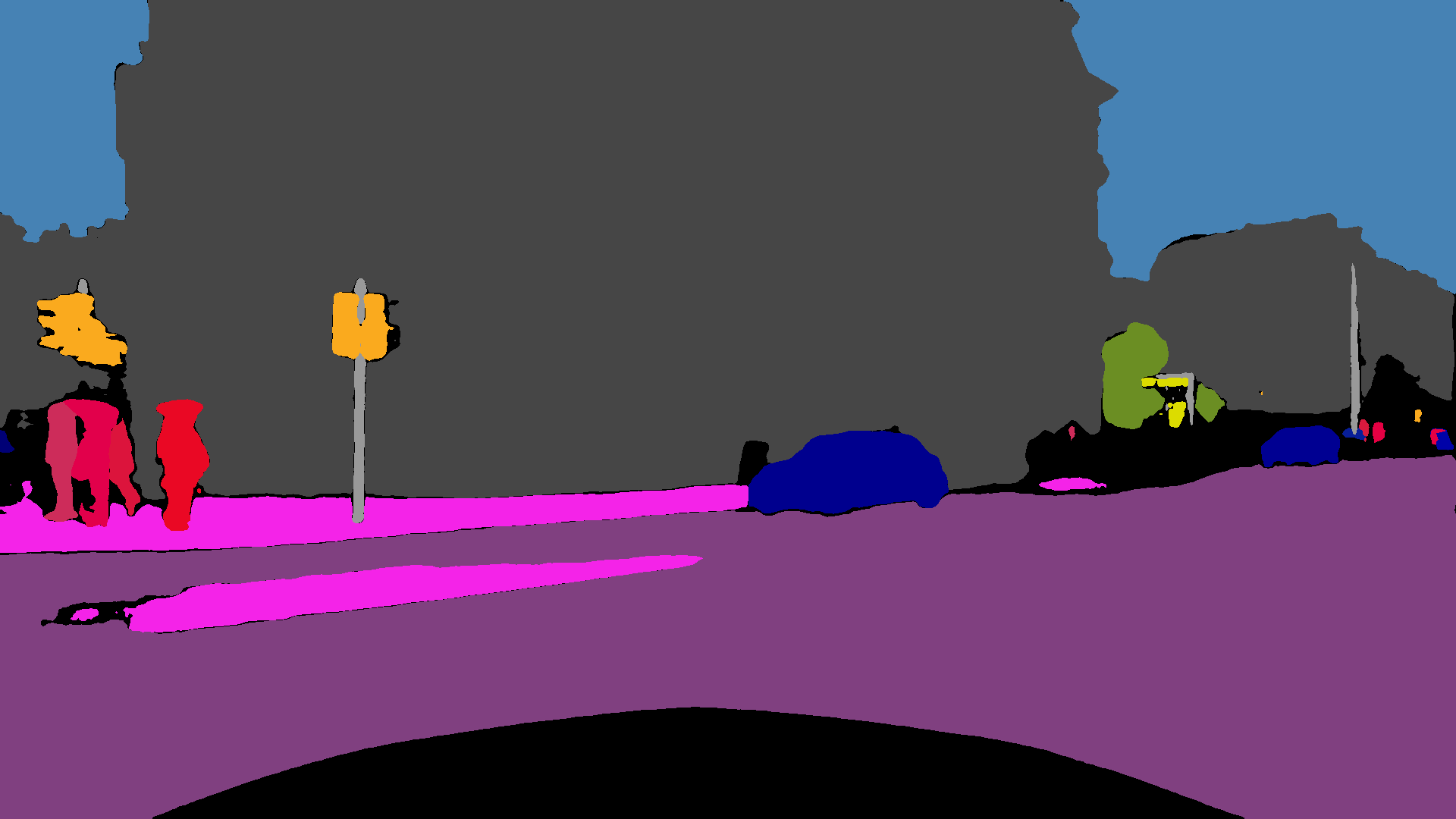} &
\includegraphics[width=0.12\textwidth]{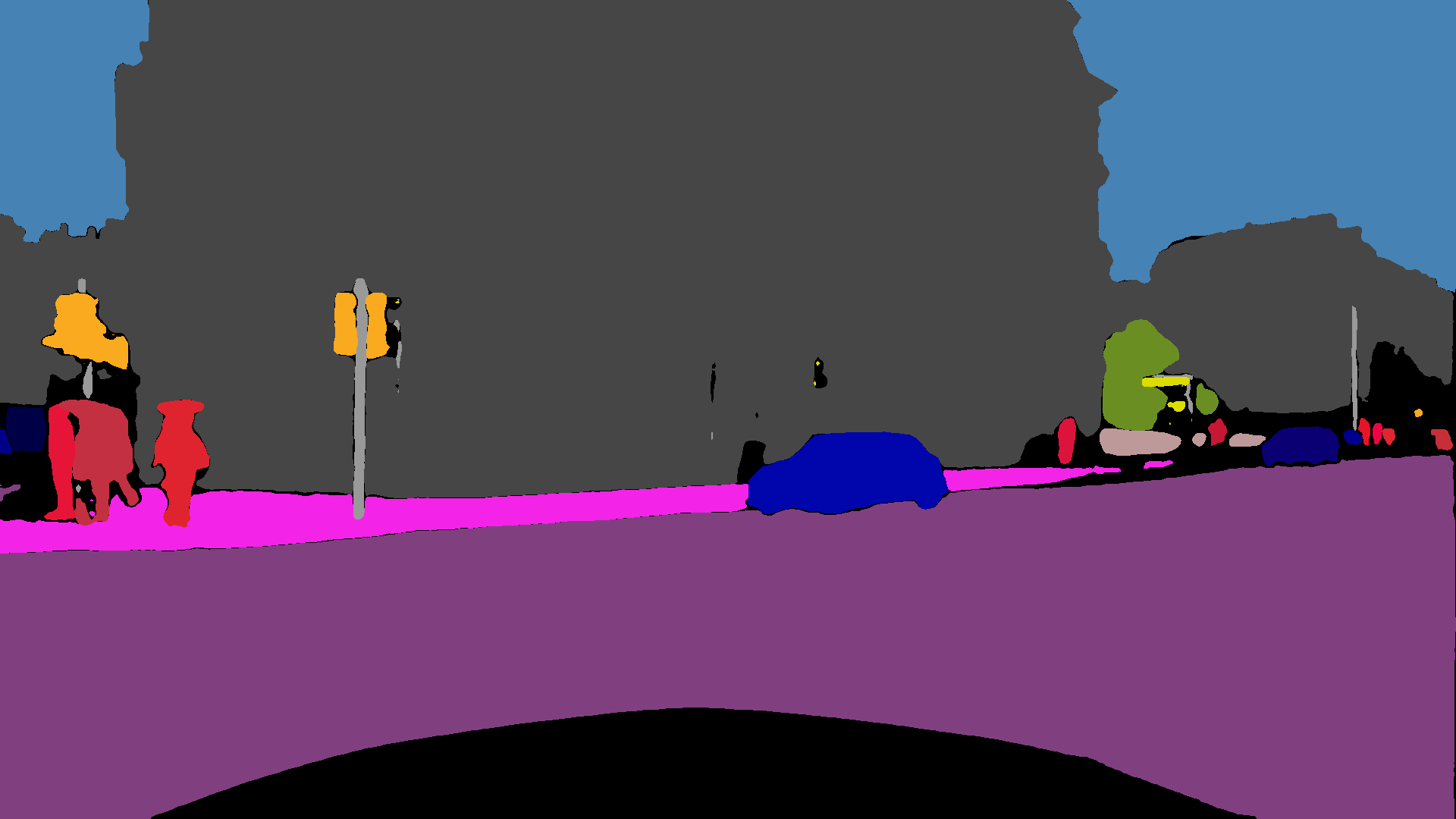} &
\includegraphics[width=0.12\textwidth]{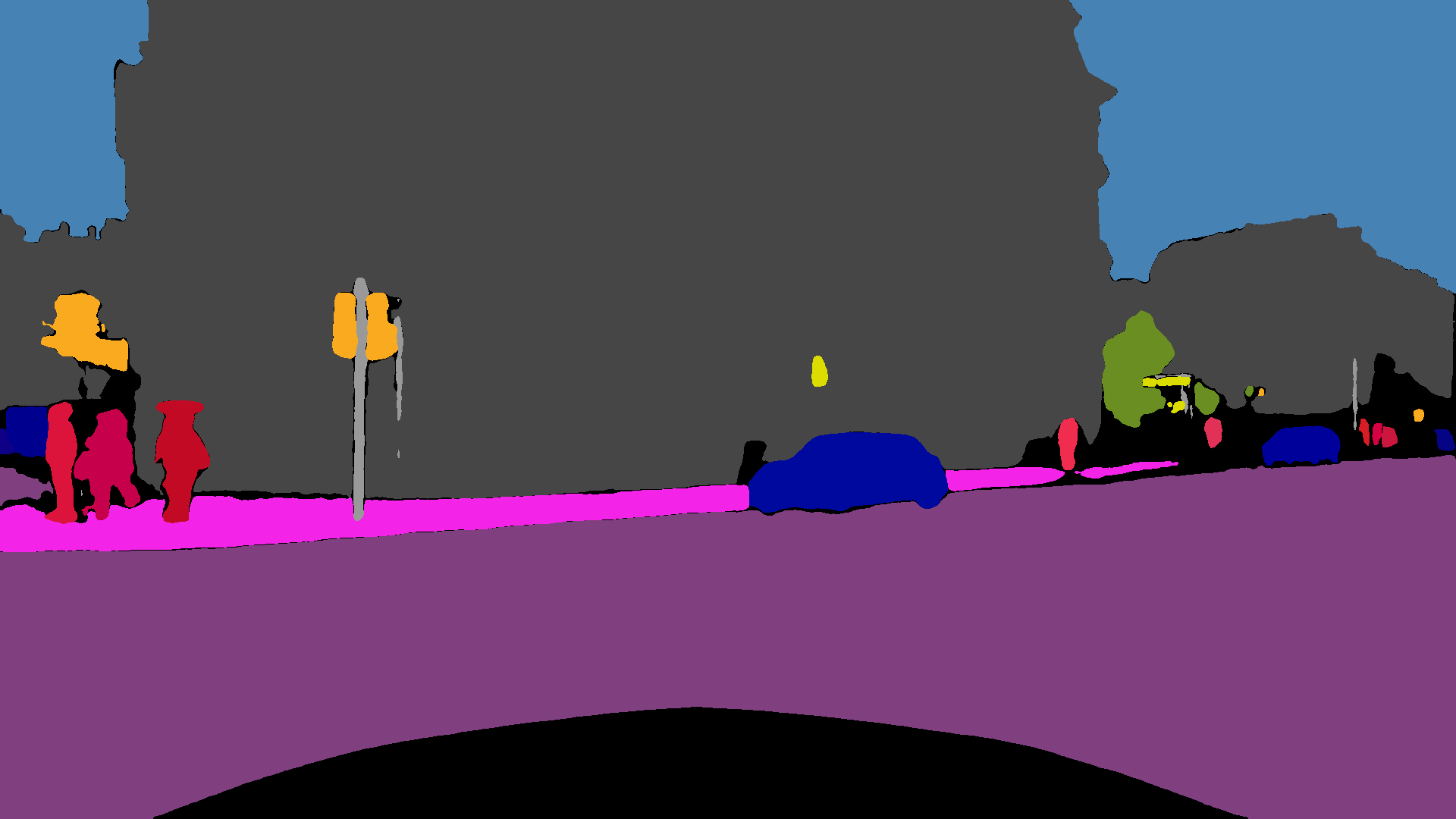} \\
\vspace{-0.1cm}

\includegraphics[width=0.12\textwidth]{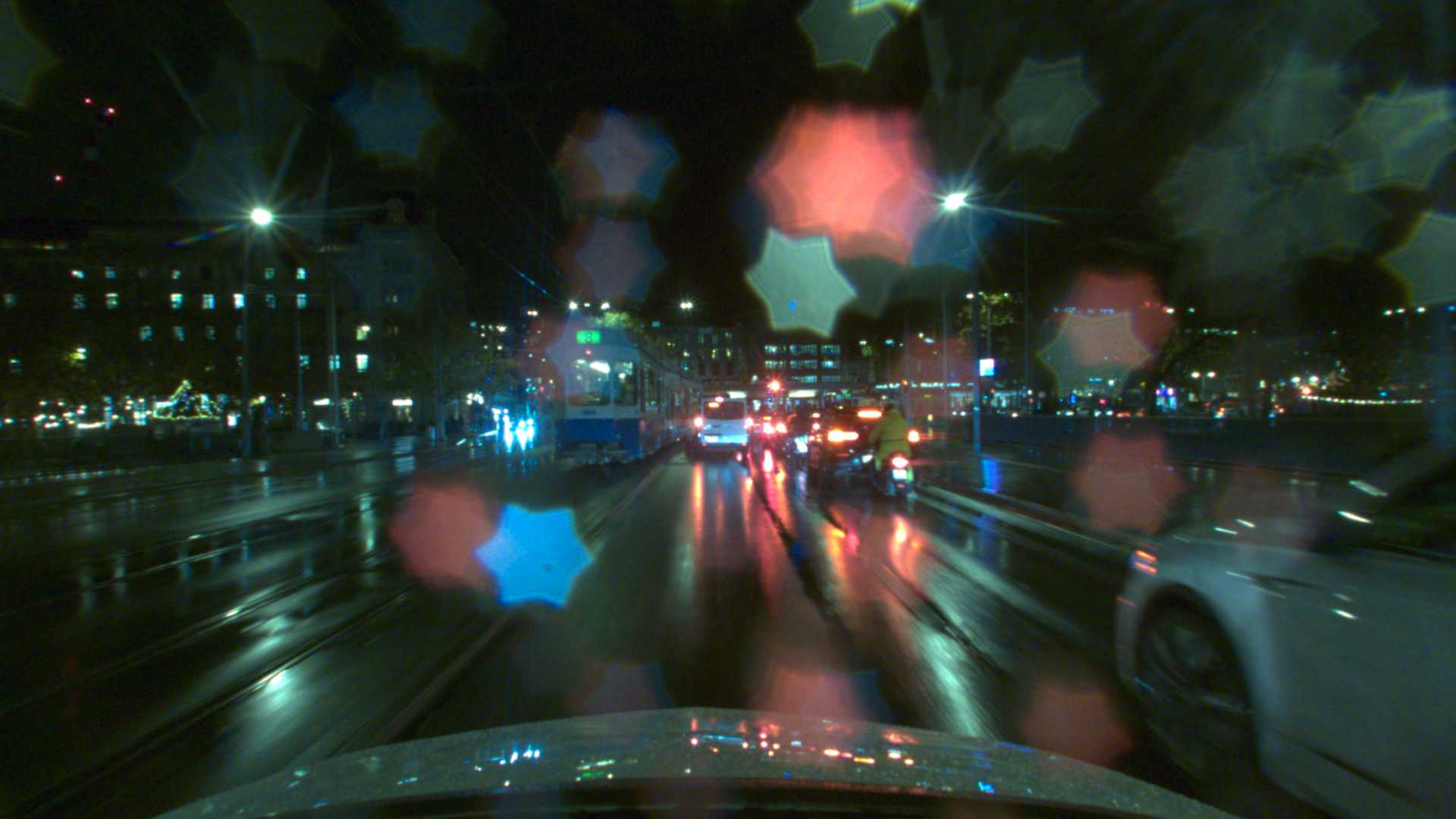} &
\includegraphics[width=0.12\textwidth]{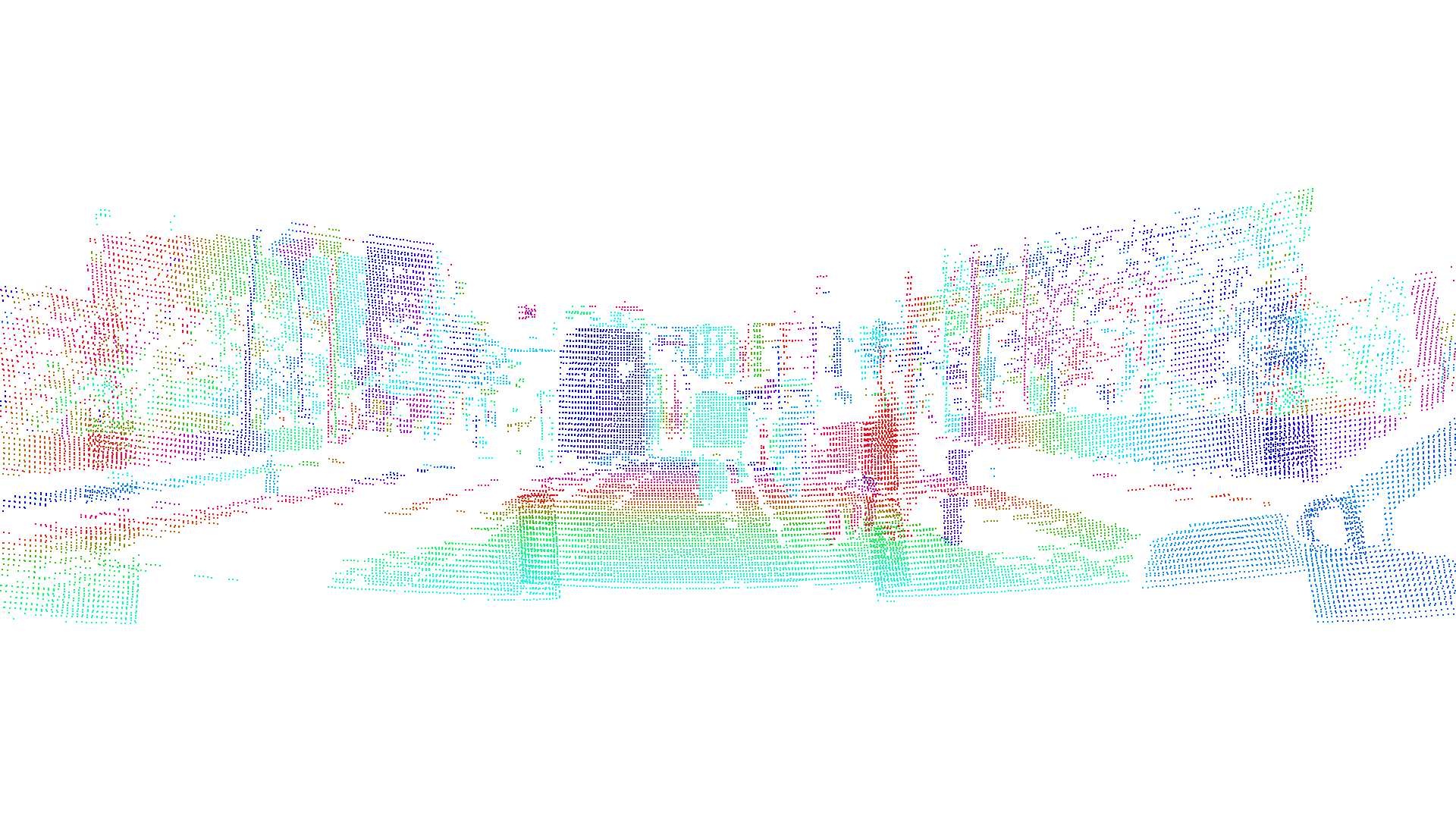} &
\includegraphics[width=0.12\textwidth]{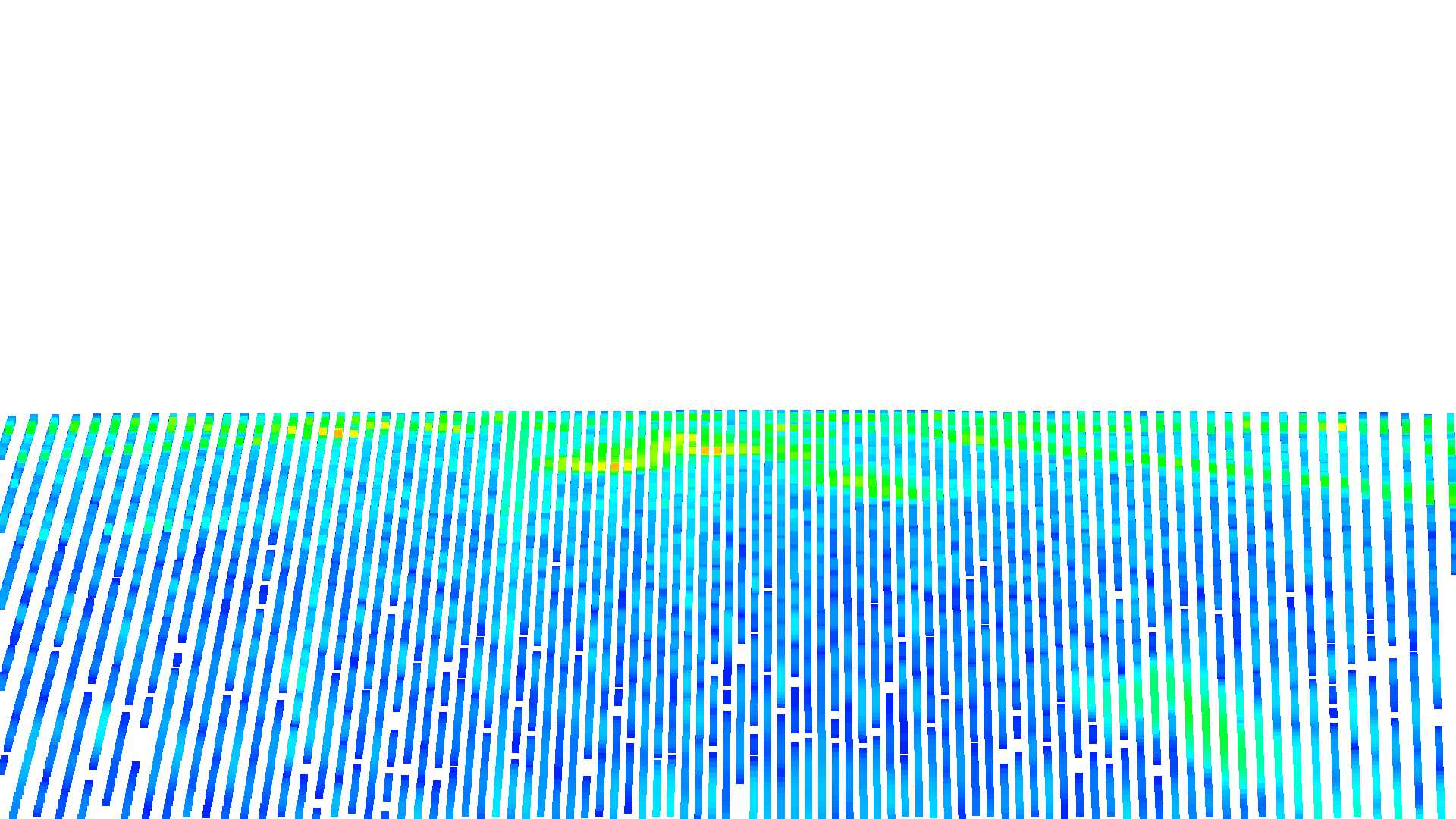} &
\includegraphics[width=0.12\textwidth]{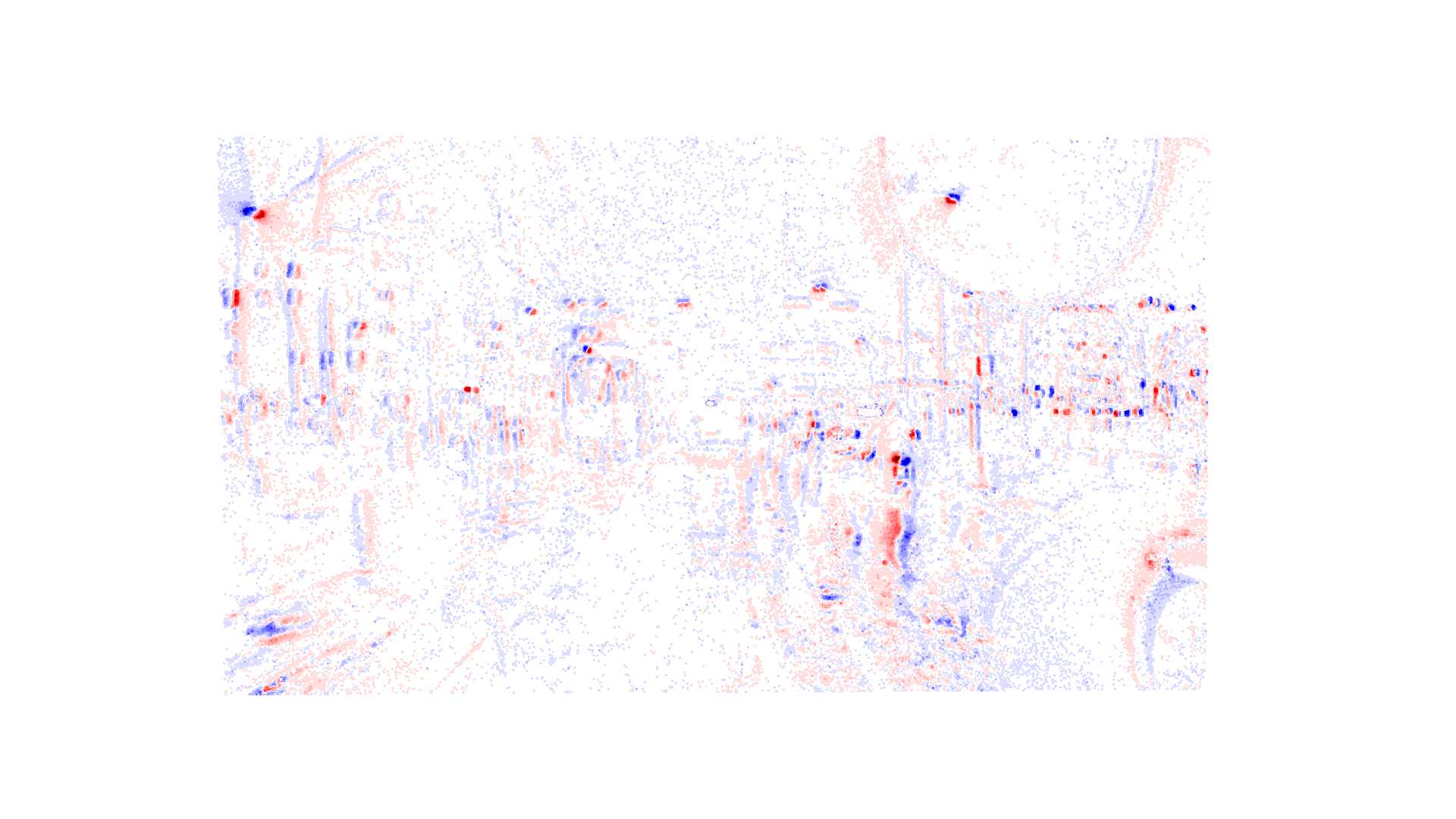} &
\includegraphics[width=0.12\textwidth]{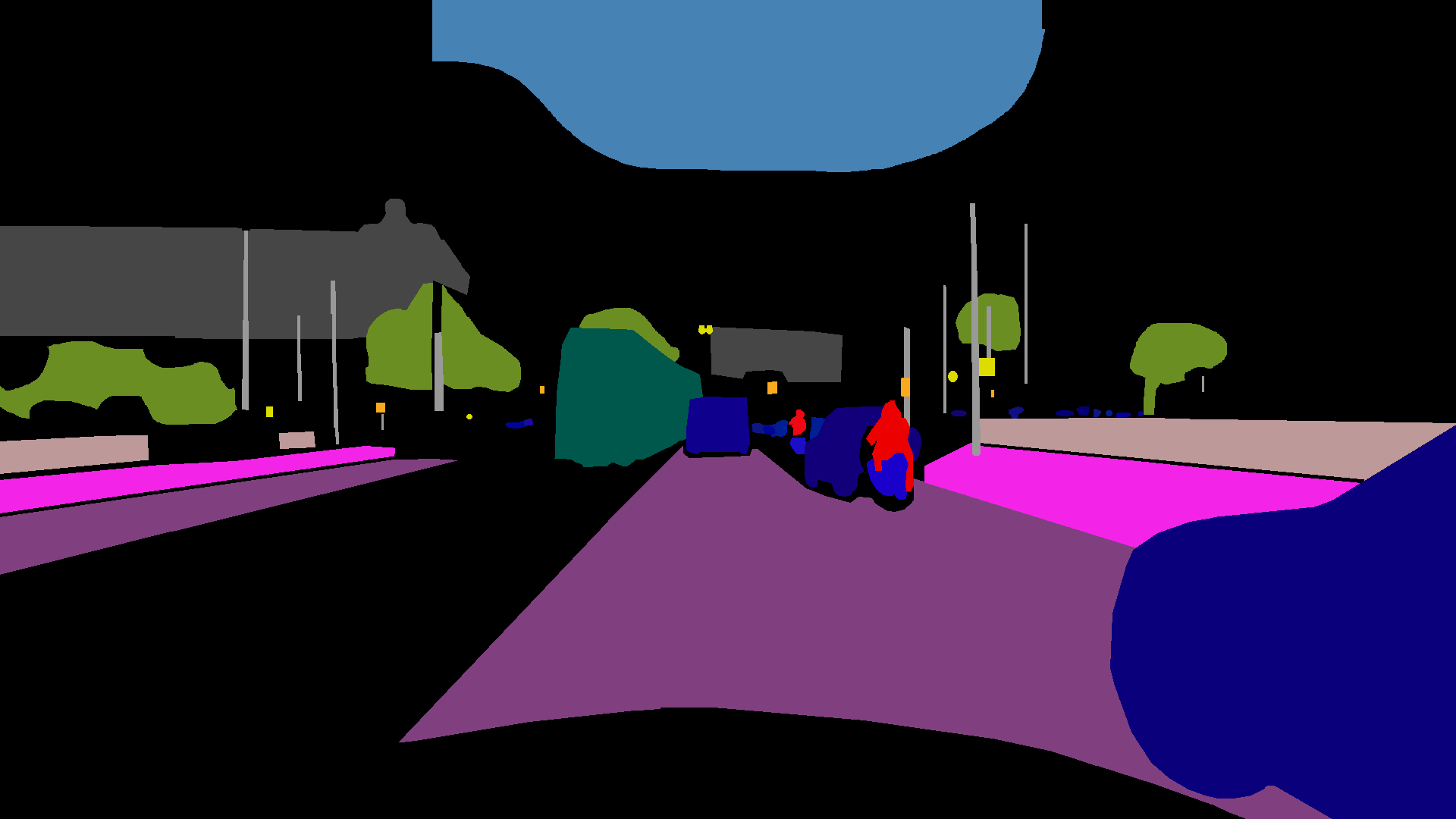} &
\includegraphics[width=0.12\textwidth]{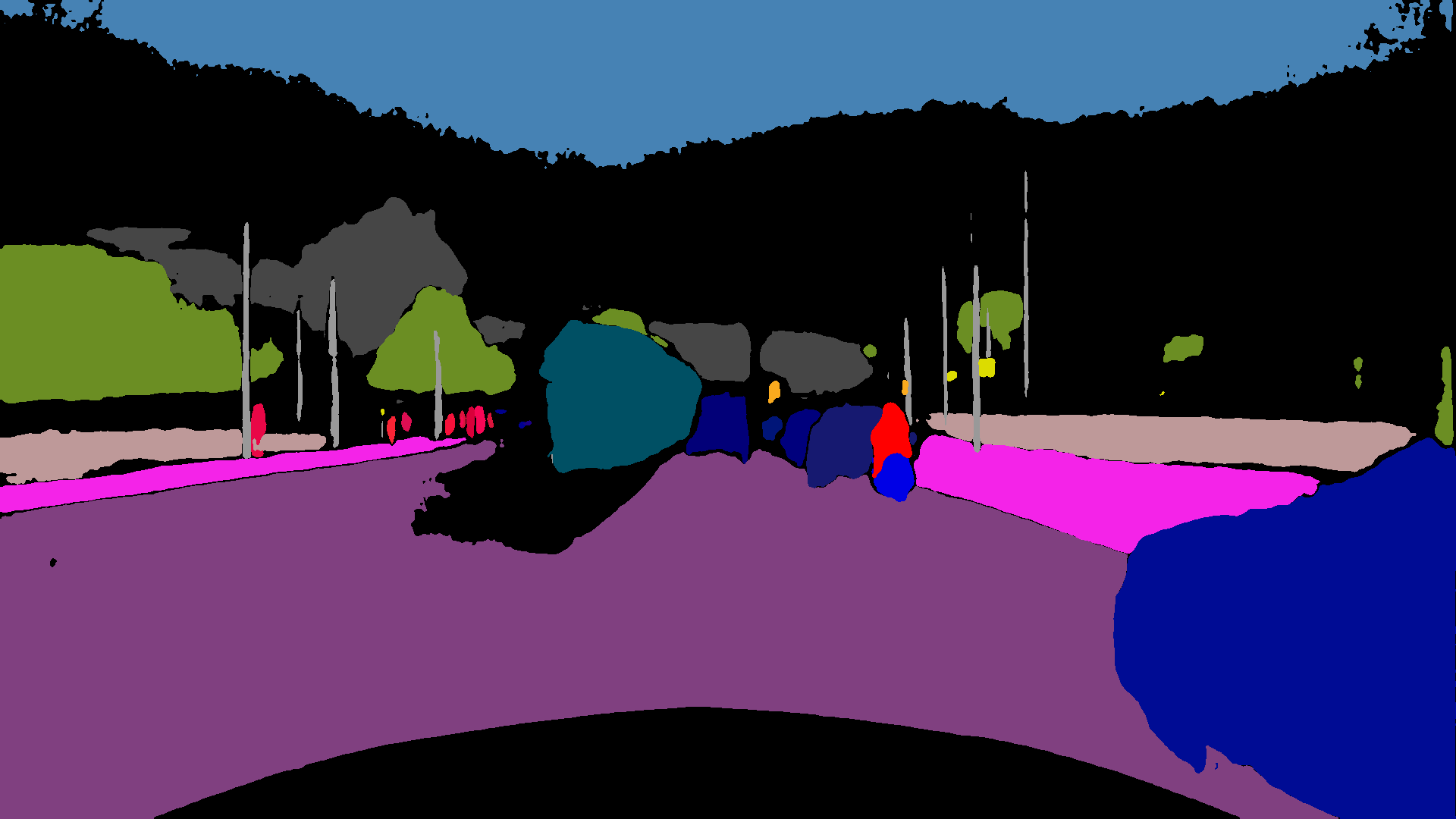} &
\includegraphics[width=0.12\textwidth]{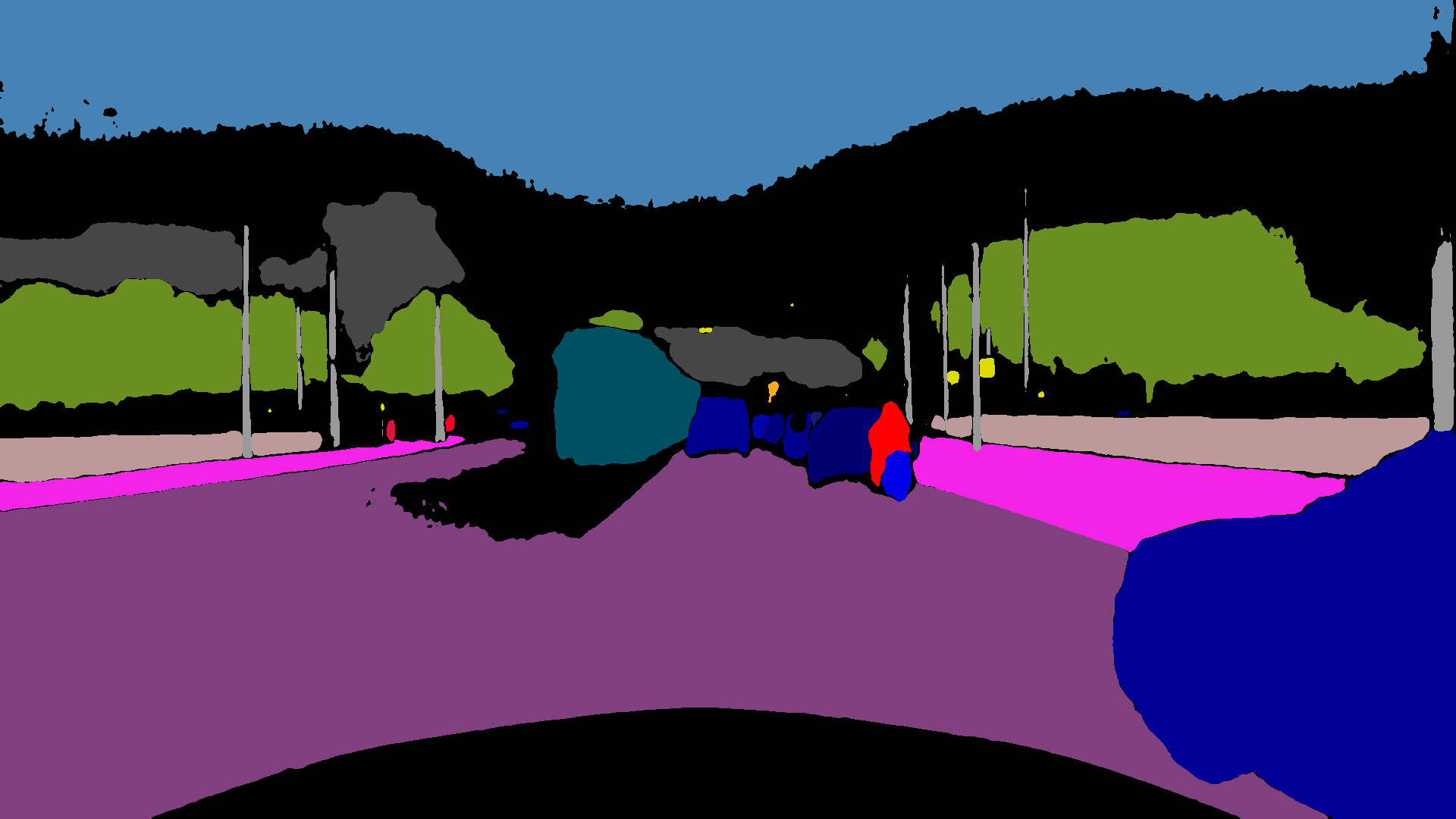} &
\includegraphics[width=0.12\textwidth]{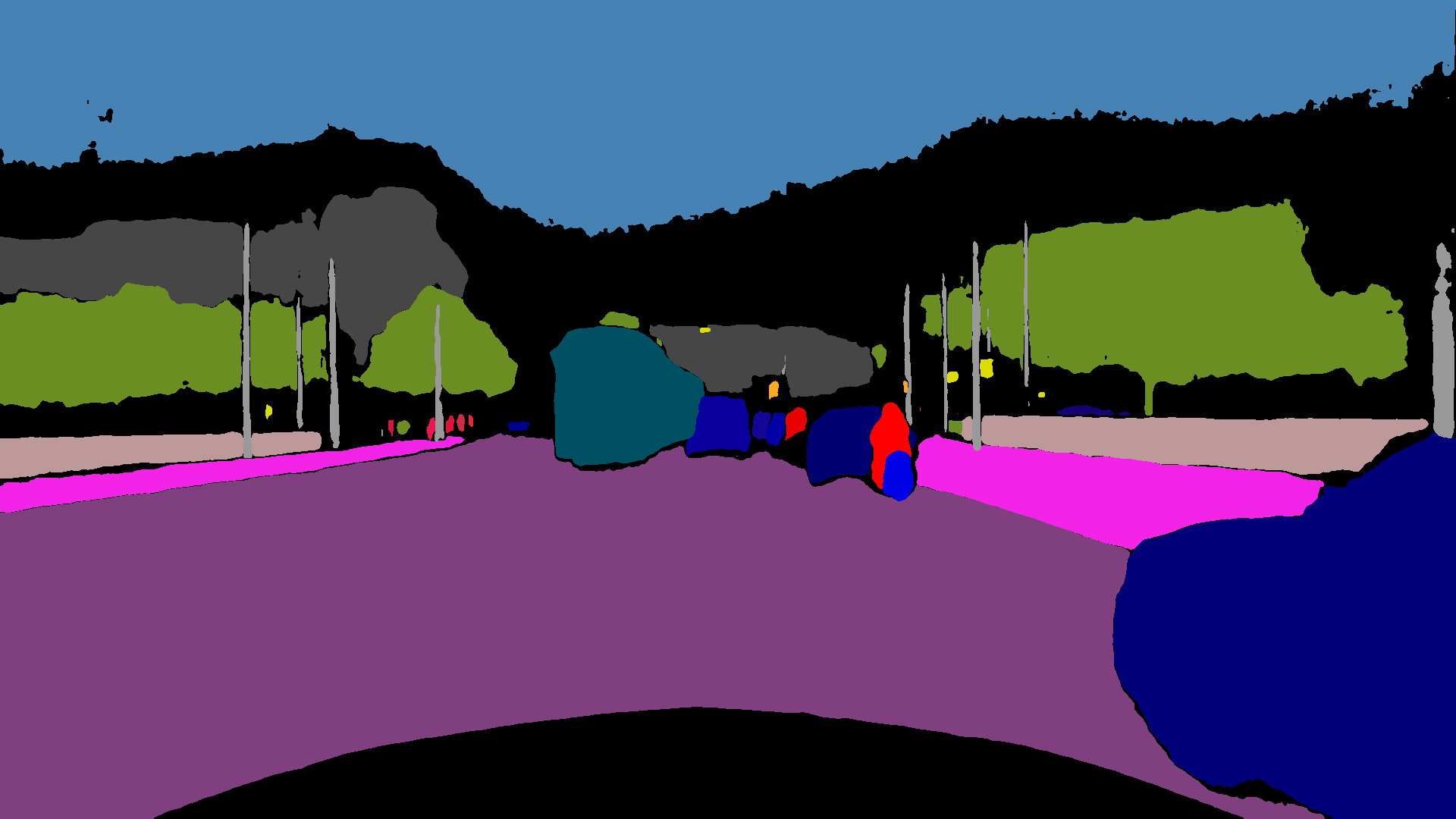} \\
\vspace{-0.1cm}

\includegraphics[width=0.12\textwidth]{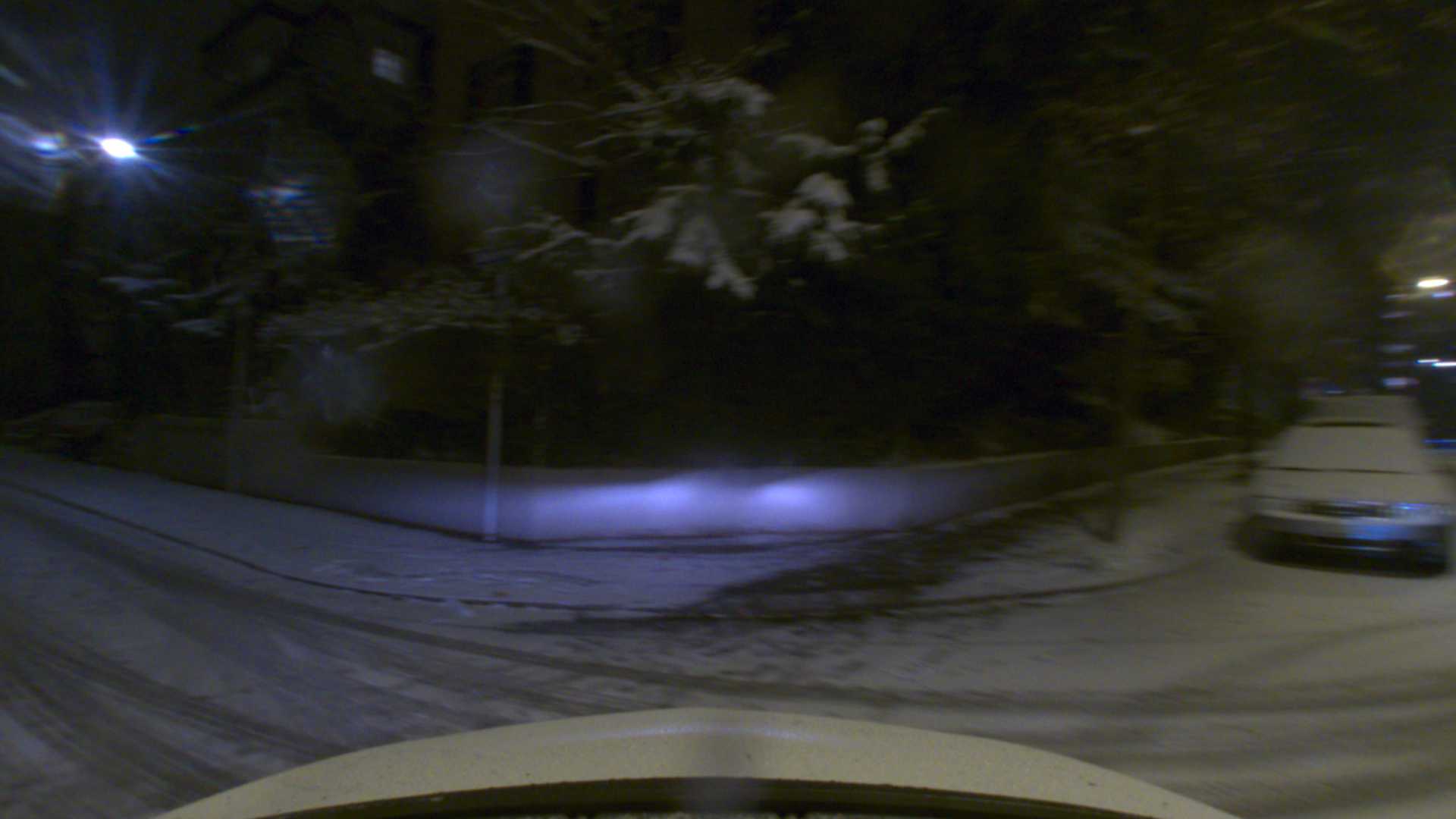} &
\includegraphics[width=0.12\textwidth]{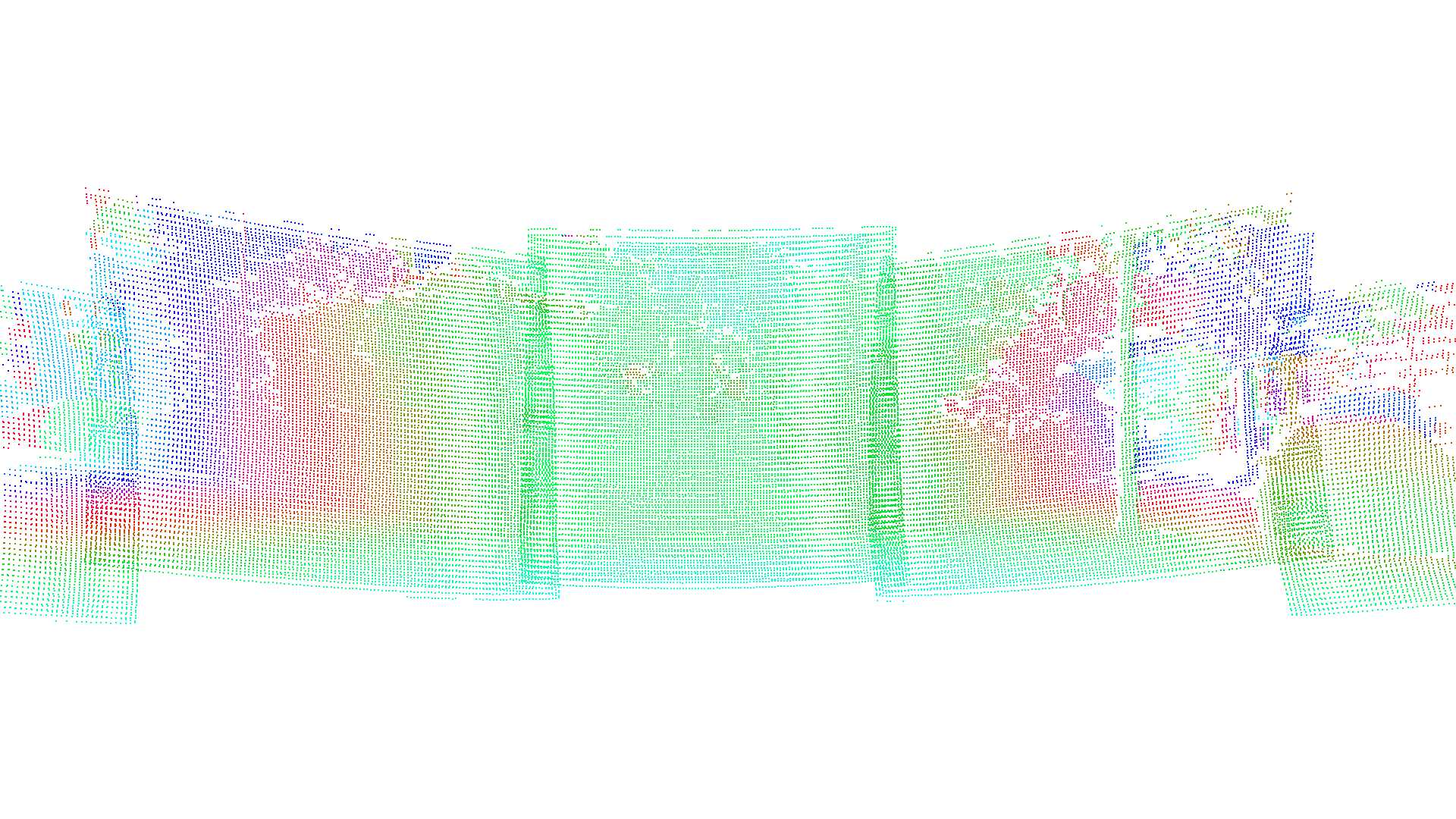} &
\includegraphics[width=0.12\textwidth]{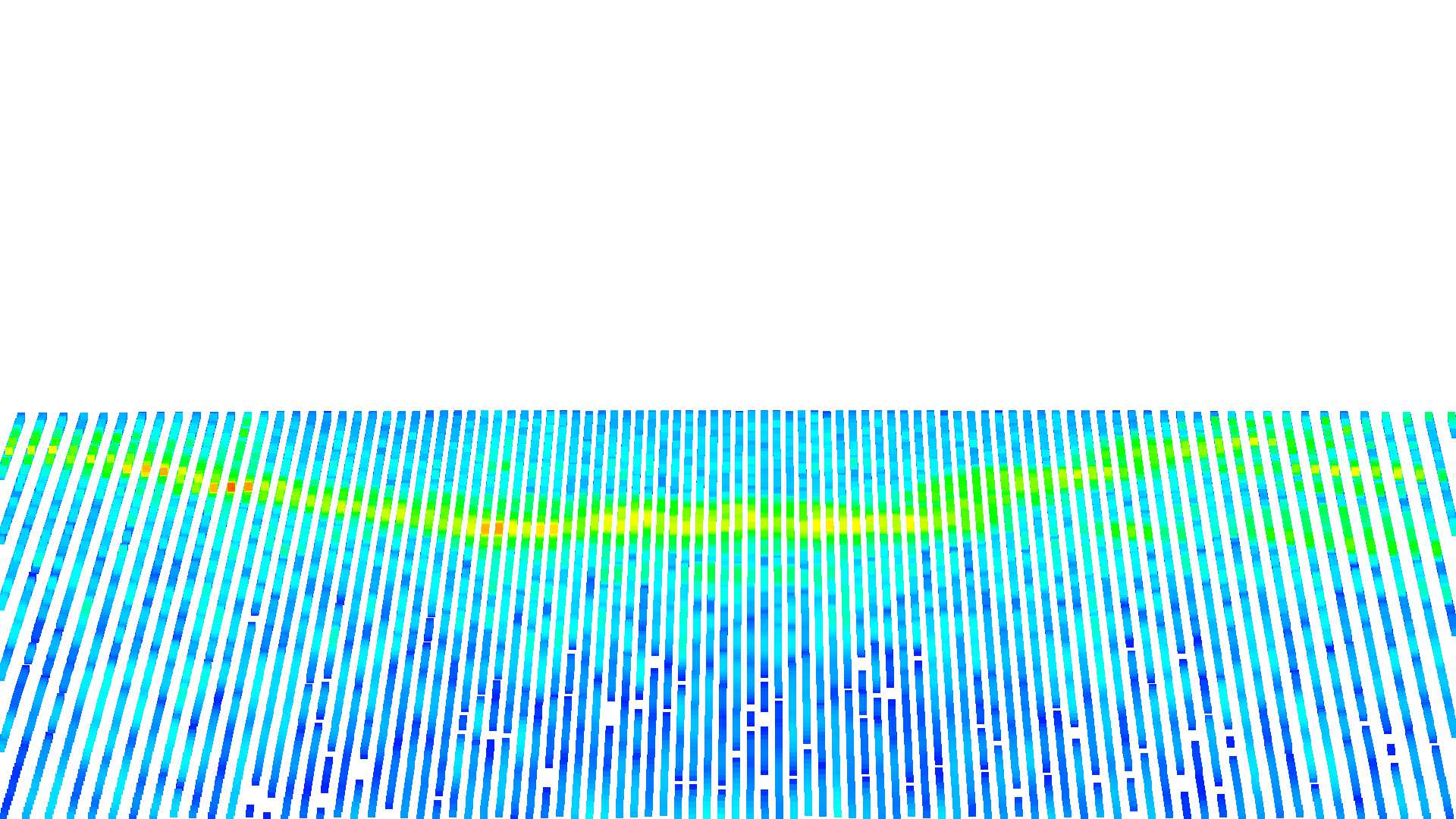} &
\includegraphics[width=0.12\textwidth]{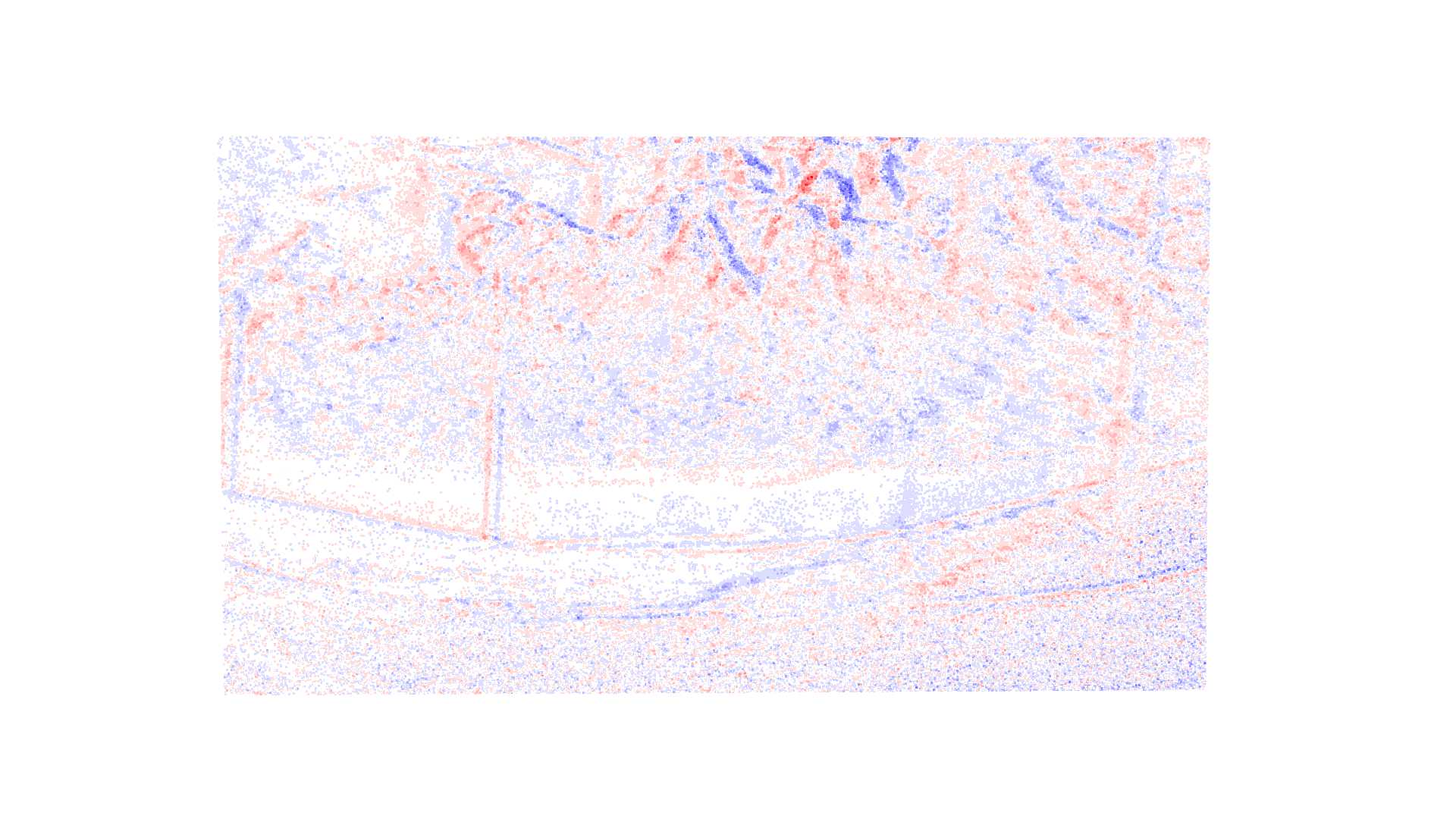} &
\includegraphics[width=0.12\textwidth]{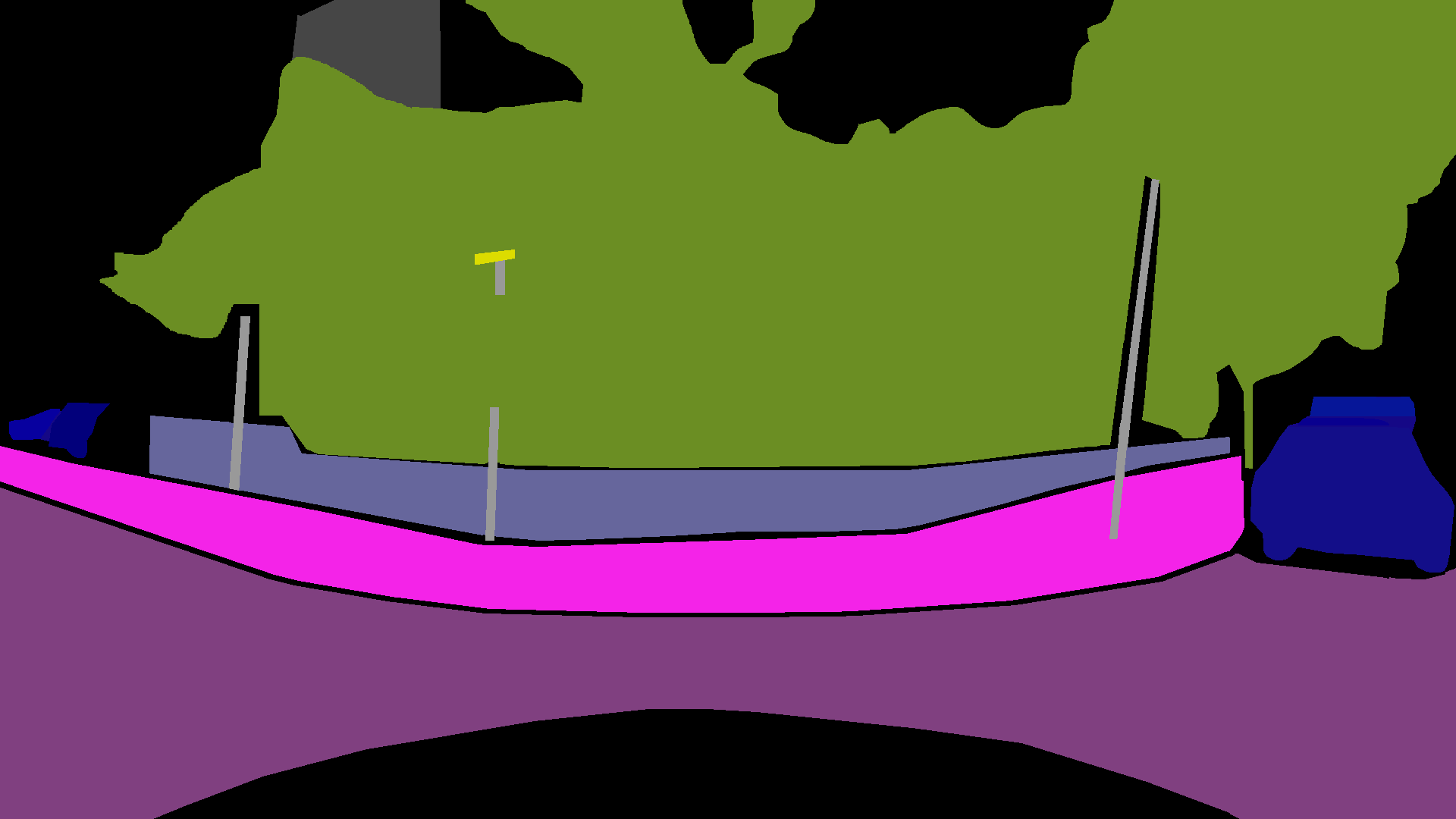} &
\includegraphics[width=0.12\textwidth]{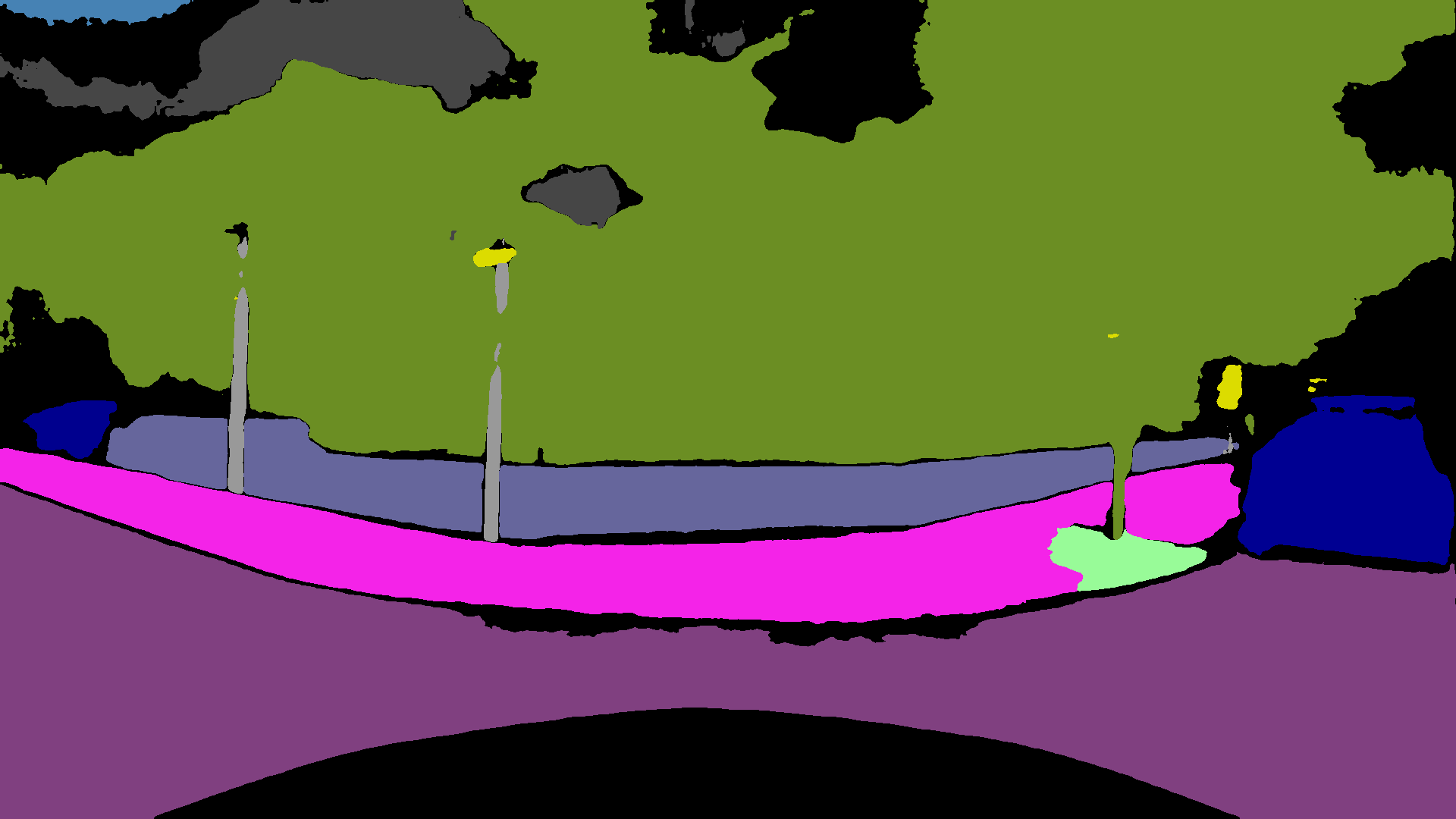} &
\includegraphics[width=0.12\textwidth]{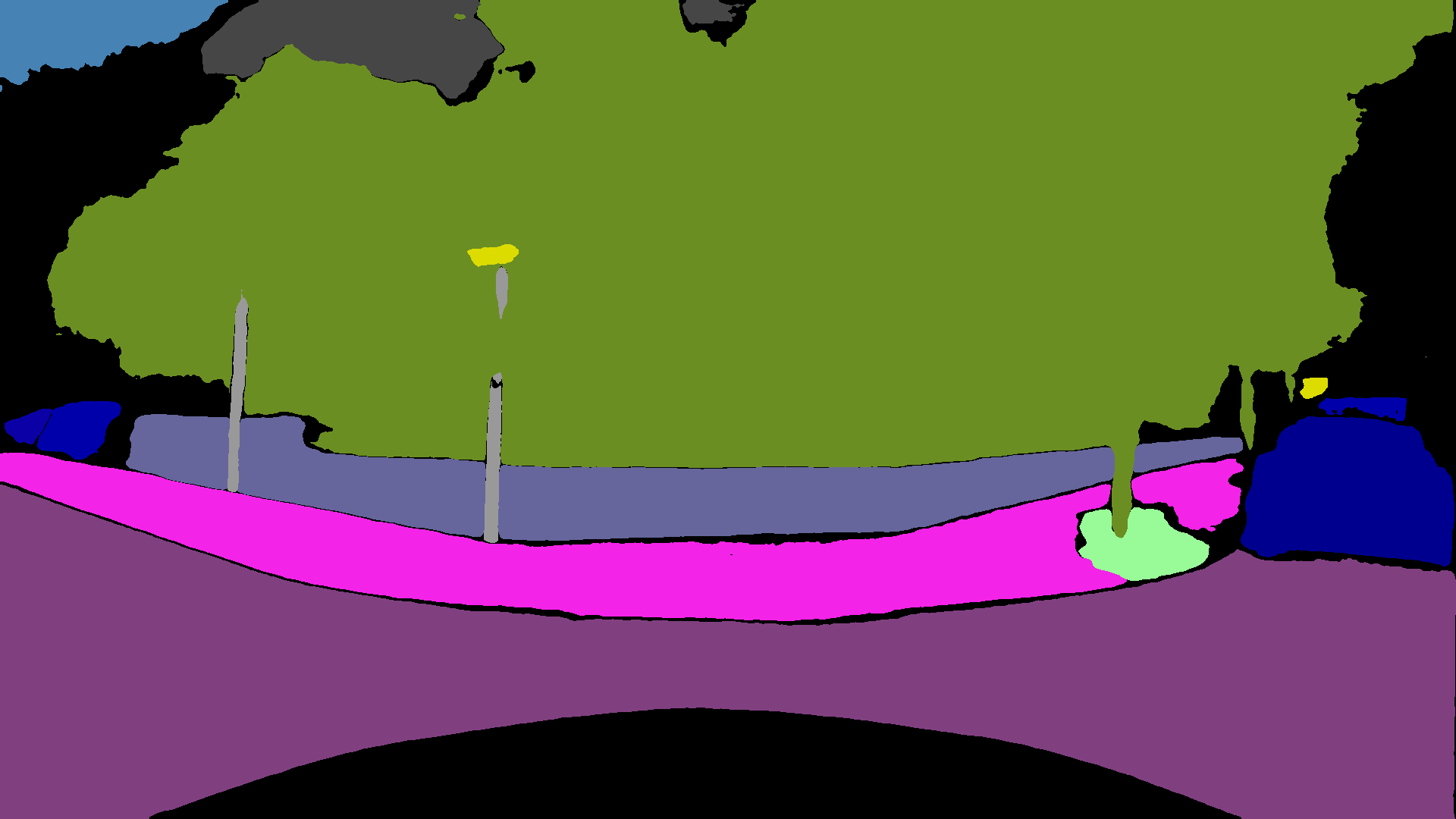} &
\includegraphics[width=0.12\textwidth]{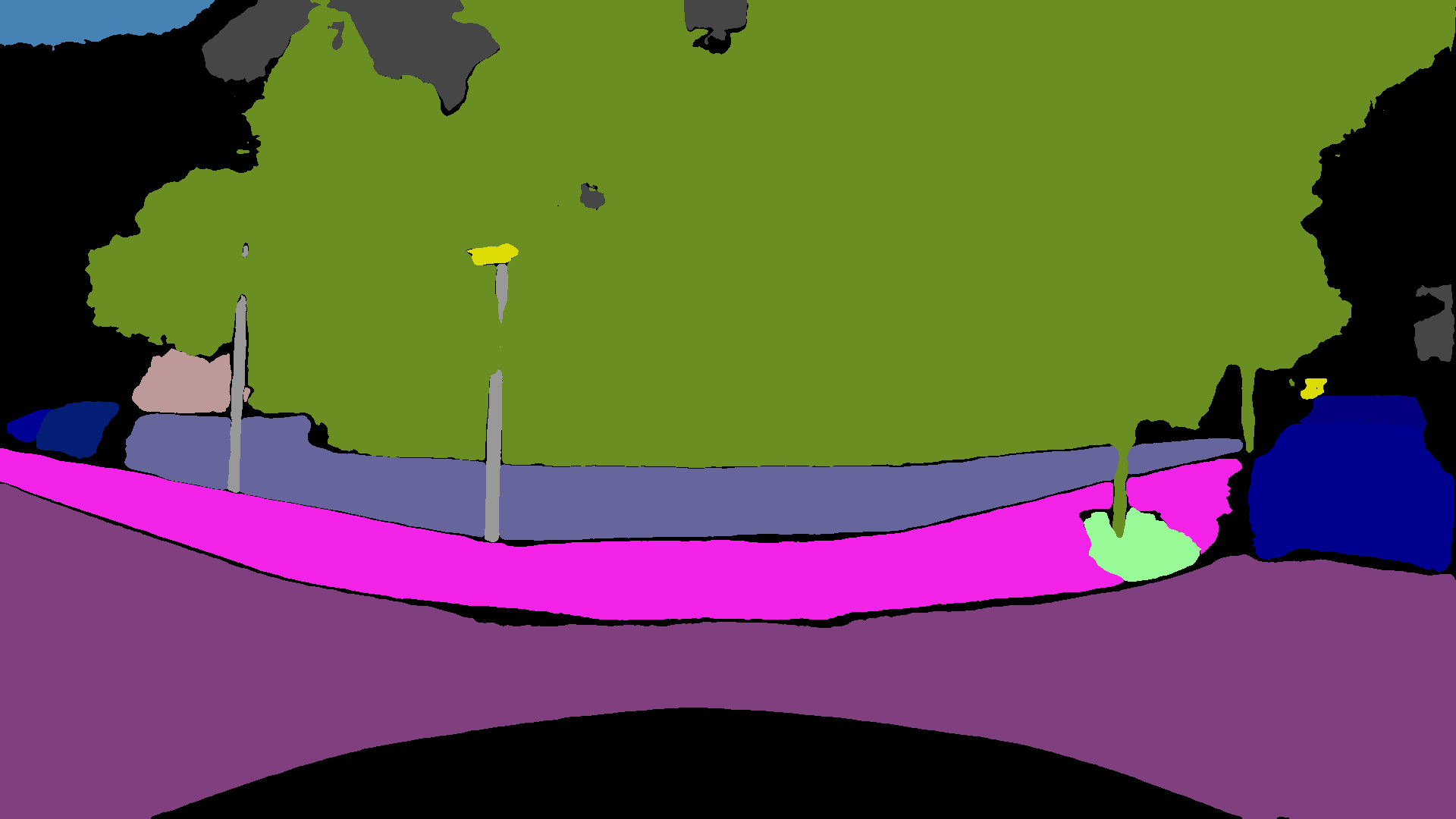} \\

\end{tabular}
\caption{Qualitative panoptic segmentation results on MUSES with visualization of the four input modalities. Best viewed on a screen at full zoom.}
\label{qual:res}
\end{figure*}

\PAR{Condition Encoding:}
In Table~\ref{tab:abl:text}, we ablate the use of different condition prompt encodings in the CT generation. Using a high-level condition description, such as simply classifying weather with a single token as e.g.\ \textit{clear} or \textit{foggy}, yields the same performance as without CAF (PQ 59.2\%). However, our proposed detailed condition prompt construction, which captures nuanced aspects of the environment (e.g., distinguishing between snow on the ground versus snow in the air), results in a significant improvement of +0.5\% in PQ, achieving a PQ of 59.7\%. This suggests that detailed condition prompts provide more fine-grained information for the fusion process, enabling more effective sensor integration based on the actual environmental context.

\PAR{Modalities:}
Table~\ref{table:abl:mod} shows the performance of our model with different combinations of input modalities. Starting with RGB-only input (55.7\% PQ), we observe consistent performance gains with each added modality: +3.0\% PQ for lidar, +0.6\% PQ for radar, and an additional +0.4\% PQ for event camera, reaching 59.7\% PQ. These results demonstrate that our approach effectively leverages all modalities, with each additional sensor contributing to performance. The flexibility of our design allows to generalize well across various input combinations.

\begin{table}[]
\centering
\caption{Ablation study on the condition prompt encodings on MUSES. 1: simple high-level condition prompt (e.g.\ [``clear'']) 2: detailed condition prompts according to \cref{sec:methods:caf}.}
\label{tab:abl:text}
\begin{tabular}{lcc}
\toprule
 & \textbf{Text Tokens}          & \textbf{PQ}$\uparrow$ \\
\hline
1  & w/o  detailed condition prompts                           & 59.2  \\
 2 (\textbf{Ours})& w/ detailed condition prompts         &  \textbf{59.7} \\
\bottomrule
\end{tabular}
\end{table}

\begin{table}
  \caption
  {Ablation for \Ours{} with different input modalities on MUSES. Each added modality improves the PQ.}
  \label{table:abl:mod}
  \smallskip
  \centering
  \setlength\tabcolsep{4pt}
  \small
  \begin{tabular*}{\linewidth}{l @{\extracolsep{\fill}} ccccl}
  \toprule
   & \textbf{RGB} & \textbf{Lidar}& \textbf{Radar} & \textbf{Events} & \textbf{PQ} $\uparrow$ \\
   \hline
   1 & \yes & \no &\no & \no &  55.7 \\
   2 & \yes & \yes &\no & \no & 58.7 (+3.0)\\
   3 & \yes & \yes &\yes & \no &  \underline{59.3} (+0.6)\\
   4 & \yes & \yes &\yes & \yes   &  \textbf{59.7} (+0.4) \\

  \bottomrule  
  \end{tabular*}
\end{table}

\subsection{CAA Weight Analysis}

We also investigate how our CAA fusion mechanism dynamically adjusts sensor modality weights in response to changing environmental conditions in scenes that were not encountered during training. \cref{fig:abl:caa:weights:all} illustrates that under clear daytime conditions, the RGB modality dominates with a weight of 68\%, capitalizing on its rich visual detail. In contrast, in foggy nighttime scenes, the RGB weight significantly decreases by -20\% to 48\%, while the weights of other modalities increase correspondingly by +8\% for lidar, +2\% for radar, and +10\% for the events. 
This substantial shift indicates that our CAA module effectively adapts to challenging conditions by decreasing reliance on less reliable sensors and boosting the weight of more robust ones per case. 
These adjustments mirror the natural strengths of each sensor modality. 
Since RGB cameras struggle in low light and in certain adverse weather conditions, the CAA fusion places greater reliance on the active sensors (lidar and radar) and event cameras, which offer a high dynamic range and excel in low-light environments. 
These findings highlight that we successfully encode environmental conditions in our CT from RGB inputs alone and adjust the fusion mechanism to prioritize the most informative sensors, thereby making semantic perception in automated driving more robust.

\subsection{Qualitative Results}
In \cref{qual:res}, we compare qualitative results of \Ours{} with competing methods. In the foggy nighttime scene (row 1), \Ours{} successfully identifies the distant car in the center. In row 2, both multimodal approaches label the person on the right, despite a droplet obscuring the RGB input. In row 3, only \Ours{} segments the more distant of the two riders waiting at the traffic light. In row 4, while all methods recognize the four vehicles on the sides, only \Ours{} achieves good instance separation without unnecessary gaps.

\addtolength{\textheight}{-1.cm}   

\section{Conclusion}

In this work, we introduced \Ours{}, a novel condition-aware multimodal fusion framework for robust semantic perception in autonomous driving. By employing a shared backbone and modality-specific feature adapters, \Ours{} efficiently aligns diverse sensor inputs into a common latent space while significantly reducing the model complexity. Our attention-based condition-aware fusion module dynamically adapts to environmental conditions guided by a Condition Token that is learned from the RGB input. This dynamic fusion enhances robustness and accuracy under challenging weather scenarios.
We demonstrated that our method outperforms competing approaches both in panoptic and semantic segmentation, setting the new state of the art on MUSES and DeLiVER. 
Extensive ablation studies demonstrate the effectiveness of our proposed modules. These advancements make \Ours{} a promising solution for enhancing semantic perception in autonomous driving and robotics, particularly under adverse environmental conditions.

{\small
\bibliographystyle{IEEEtran}
\bibliography{IEEEabrv,main}
}

\end{document}